\documentclass{article} % ICLR 2027 official shell uses article with the ICLR style.
\usepackage{05_paper/arxiv_preprint,times}

\usepackage{amsmath,amsfonts,bm}

\def\figref#1{figure~\ref{#1}}
\def\eqref#1{equation~\ref{#1}}
\def\1{\bm{1}}

\DeclareMathAlphabet{\mathsfit}{\encodingdefault}{\sfdefault}{m}{sl}
\SetMathAlphabet{\mathsfit}{bold}{\encodingdefault}{\sfdefault}{bx}{n}

\usepackage{hyperref}
\usepackage{url}
\usepackage{graphicx}
\usepackage{wrapfig}
\usepackage{booktabs}
\usepackage{placeins}
\usepackage{tabularx}
\usepackage{longtable}
\usepackage{pdflscape}
\usepackage{rotating}
\usepackage{multicol}
\usepackage{array}
\usepackage{xcolor}
\usepackage{tcolorbox}
\setcitestyle{yysep={,}}
\definecolor{citationgreen}{HTML}{6F9B86}
\definecolor{rqblue}{HTML}{78A9C9}
\definecolor{FindingBackground}{HTML}{F7F4F9}
\definecolor{FindingAccent}{HTML}{C6B4D8}
\definecolor{FindingTag}{HTML}{B49BCB}
\definecolor{FindingHeadline}{HTML}{35303A}
\definecolor{FindingSupport}{HTML}{6A6273}
\definecolor{FindingShadow}{HTML}{EAE3EF}
\definecolor{AppendixPink}{HTML}{C79AA5}
\definecolor{PaperLinkPurple}{HTML}{6E5A8A}
\definecolor{FilteringRuleBackground}{HTML}{FAF7FB}
\definecolor{FilteringRuleTitle}{HTML}{6E6177}
\definecolor{FilteringRuleHeading}{HTML}{514957}
\definecolor{FilteringRuleBody}{HTML}{3F3B42}
\definecolor{FilteringRuleDivider}{HTML}{EEE7F1}
\definecolor{DefinitionBackground}{HTML}{FAF8FB}
\definecolor{DefinitionOutline}{HTML}{EAE4EF}
\definecolor{DefinitionHeading}{HTML}{3B3640}
\definecolor{DefinitionBody}{HTML}{625B69}
\definecolor{RoadmapConstruction}{HTML}{F4F8FB}
\definecolor{RoadmapComposition}{HTML}{F4F8F6}
\definecolor{RoadmapEvaluation}{HTML}{FAF6F7}
\definecolor{RoadmapSupporting}{HTML}{F8F5FA}
\definecolor{RoadmapHeading}{HTML}{38353B}
\definecolor{RoadmapText}{HTML}{4E4A52}
\definecolor{MainTableHeader}{HTML}{EEF3F5}
\newlength{\MainTableHeaderWidth}
\newcommand{\MainTableHeaderBackground}{%
  \tikz[overlay,baseline]{%
    \fill[MainTableHeader,rounded corners=1.5mm]
      (0pt,-2.25pt) rectangle (\MainTableHeaderWidth,7.75pt);%
  }%
}
\hypersetup{
  colorlinks=true,
  citecolor=PaperLinkPurple,
  linkcolor=PaperLinkPurple,
  urlcolor=PaperLinkPurple
}
\newcommand{\jpegext}{\mbox{\texttt{.\kern-0.34em{}jpeg}}} % Tight code-font rendering avoids a visual ". jpeg" gap.
\newcommand{\catname}[1]{\texttt{#1}}
\newcommand{\appref}[1]{%
  \hyperref[#1]{\textcolor{PaperLinkPurple}{Appendix~\ref*{#1}}}%
}
\providecommand{\figref}[1]{}
\renewcommand{\figref}[1]{%
  \hyperref[#1]{\textcolor{PaperLinkPurple}{Fig.~\ref*{#1}}}%
}
\newcommand{\figpanelref}[2]{%
  \hyperref[#1]{\textcolor{PaperLinkPurple}{Fig.~\ref*{#1}(#2)}}%
}
\newcommand{\appfigref}[1]{%
  \hyperref[#1]{\textcolor{PaperLinkPurple}{Appendix Figure~\ref*{#1}}}%
}
\providecommand{\tabref}[1]{}
\renewcommand{\tabref}[1]{%
  \hyperref[#1]{\textcolor{PaperLinkPurple}{Tab.~\ref*{#1}}}%
}
\newtcbox{\findingpill}{%
  on line,
  nobeforeafter,
  colback=FindingTag,
  colframe=FindingTag,
  boxrule=0pt,
  arc=1.7mm,
  outer arc=1.7mm,
  boxsep=0pt,
  left=1.5mm,
  right=1.5mm,
  top=0.45mm,
  bottom=0.45mm,
  fontupper=\bfseries\footnotesize,
  colupper=white
}
\tcbset{%
  findingcard/.style={%
    colback=FindingBackground,
    colframe=FindingBackground,
    boxrule=0pt,
    arc=3.5mm,
    outer arc=3.5mm,
    boxsep=0pt,
    left=5mm,
    right=4.5mm,
    top=2.6mm,
    bottom=2.6mm,
    before skip=4pt,
    after skip=4pt,
    fontupper=\normalfont,
    coltext=FindingSupport
  }
}
\tcbset{%
  filteringrulescard/.style={%
    colback=FilteringRuleBackground,
    colframe=FilteringRuleBackground,
    boxrule=0pt,
    arc=2.7mm,
    outer arc=2.7mm,
    boxsep=0pt,
    left=4mm,
    right=4mm,
    top=2.7mm,
    bottom=2.7mm,
    before skip=4pt,
    after skip=4pt,
    fontupper=\normalfont,
    coltext=FilteringRuleBody
  }
}
\newsavebox{\findingcardsave}
\newsavebox{\findingpillsave}
\newlength{\findinglabelwidth}
\newlength{\findingaccentheight}
\newcommand{\findingbox}[3]{%
  \par\smallskip
  \noindent\sbox{\findingcardsave}{%
    \begin{minipage}{\dimexpr\linewidth-2mm\relax}
      \begin{tcolorbox}[findingcard,before skip=0pt,after skip=0pt]
        \noindent
        \sbox{\findingpillsave}{\findingpill{FINDING #1}}%
        \setlength{\findinglabelwidth}{\wd\findingpillsave}%
        \renewcommand{\tabularxcolumn}[1]{m{##1}}%
        \begin{tabularx}{\linewidth}{@{}>{\centering\arraybackslash}m{\findinglabelwidth}@{\hspace{1mm}}>{\raggedright\arraybackslash}X@{}}
          \usebox{\findingpillsave} & {\color{FindingHeadline}\bfseries #2}
        \end{tabularx}
        \par\vspace{0.8mm}%
        {\color{FindingSupport}#3}
      \end{tcolorbox}
    \end{minipage}%
  }%
  \setlength{\findingaccentheight}{\dimexpr\ht\findingcardsave+\dp\findingcardsave\relax}%
  \setlength{\findingaccentheight}{0.66\findingaccentheight}%
  \begin{tikzpicture}
    \node[anchor=south west,inner sep=0pt,outer sep=0pt,draw=none,
      fill=FindingShadow,rounded corners=3.5mm,
      minimum width=\wd\findingcardsave,
      minimum height=\dimexpr\ht\findingcardsave+\dp\findingcardsave\relax]
      at (0.3mm,-0.6mm) {};
    \node[anchor=south west,inner sep=0pt,outer sep=0pt] (findingmain) at (0,0) {\usebox{\findingcardsave}};
    \node[anchor=center,inner sep=0pt,outer sep=0pt,draw=none,
      fill=FindingAccent,rounded corners=1.1pt,
      minimum width=2.2pt,minimum height=\findingaccentheight]
      at ([xshift=2.8mm]findingmain.west) {};
  \end{tikzpicture}%
  \par\smallskip
}
\newcommand{\rqbox}[2]{%
  \par\smallskip
  \noindent{\color{rqblue}\bfseries #1. #2}\par
  \par\smallskip
}
\tcbset{%
  definitioncardstyle/.style={%
    colback=DefinitionBackground,
    colframe=DefinitionOutline,
    boxrule=0.4pt,
    arc=3mm,
    outer arc=3mm,
    boxsep=0pt,
    left=3.8mm,
    right=3.8mm,
    top=2.7mm,
    bottom=2.7mm,
    before skip=4pt,
    after skip=4pt,
    fontupper=\normalfont,
    coltext=DefinitionBody
  },
  roadmapgroup/.style={%
    boxrule=0pt,
    arc=2.7mm,
    outer arc=2.7mm,
    boxsep=0pt,
    left=2.8mm,
    right=2.8mm,
    top=1.7mm,
    bottom=1.5mm,
    before skip=3pt,
    after skip=3pt,
    fontupper=\small,
    coltext=RoadmapText
  }
}
\newenvironment{definitioncard}[1]{%
  \par\smallskip
  \begin{tcolorbox}[definitioncardstyle]
    \noindent{\color{DefinitionHeading}\bfseries Definition 1. #1}\par\vspace{0.8mm}%
    \color{DefinitionBody}\ignorespaces
}{%
  \end{tcolorbox}
  \par\smallskip
}

\title{What Remains OOD? OODConceptBench for Open-World Concept Recognition in Vision--Language Models}

\author{
\textbf{Zhiran Luo}$^{1}$,
\textbf{Zhiyong Yang}$^{1}$,
\textbf{Zitai Wang}$^{2}$,
\textbf{Sicong Li}$^{3,4}$,
\textbf{Qianqian Xu}$^{2,5}$,
\textbf{Qingming Huang}$^{1,2}$\\[0.5em]
\small $^{1}$School of Computer Science and Technology,
University of Chinese Academy of Sciences, Beijing, China\\
\small $^{2}$Institute of Computing Technology,
Chinese Academy of Sciences, Beijing, China\\
\small $^{3}$Institute of Information Engineering,
Chinese Academy of Sciences, Beijing, China\\
\small $^{4}$School of Cyber Security,
University of Chinese Academy of Sciences, Beijing, China\\
\small $^{5}$Beijing Academy of Artificial Intelligence,
Beijing, China
}

\begin{document}

\maketitle

\setlength{\textfloatsep}{6pt}
\setlength{\floatsep}{6pt}
\setlength{\intextsep}{6pt}

\begin{abstract}
 In the era of large-scale pretrained models, zero-shot generalization has increasingly blurred the boundary between known and unknown concepts. However, current evaluations do not directly reveal which open-world concepts different vision--language models can recognize. To study this problem, we introduce OODConceptBench, a hierarchical benchmark for open-world concept recognition on image--concept pairs. The full benchmark set (L1) contains 30,752 images, and the harder subset (L2) contains 14,285 images. The benchmark covers 560 concept phrases across seven open-world categories. We evaluate 37 vision--language models across release year, model family, and variant. We obtain three main findings. First, target-concept recognition is lower on L2 than on L1 across all 37 evaluated models. Second, later-released models do not necessarily achieve stronger open-world concept recognition. Third, an increase in overall recognition does not imply consistent improvement across all benchmark categories. Together, these results show that OOD remains a challenge even in the era of large-scale vision--language pretraining.
\end{abstract}

\section{Introduction}

\begin{figure}[t]
    \centering
    \includegraphics[width=0.98\textwidth,keepaspectratio]{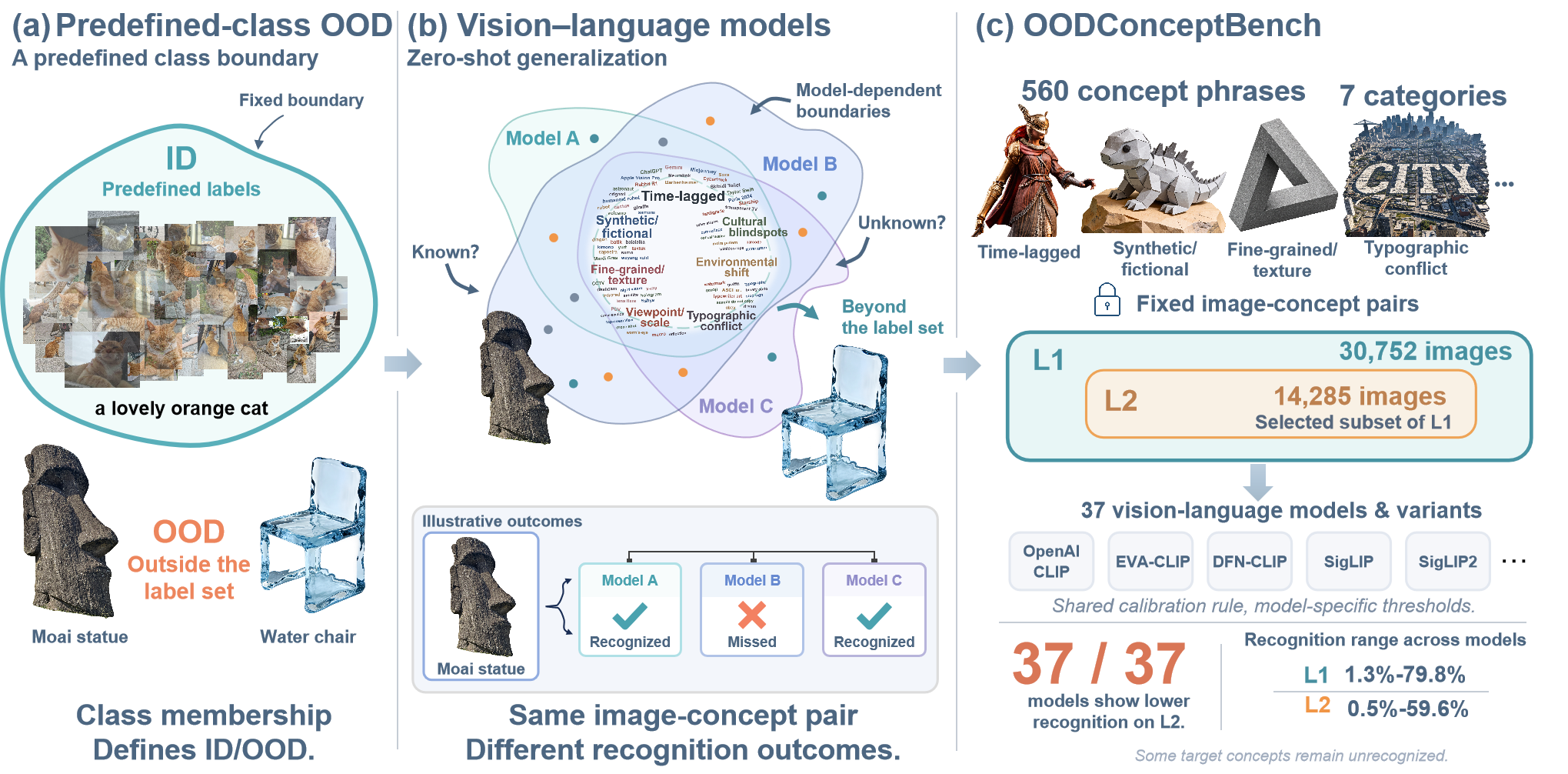}
    \caption{From predefined-class OOD to model-specific concept recognition. OODConceptBench fixes image--concept pairs and compares recognition across models using a shared calibration rule with model-specific recognition thresholds.}
    \label{fig:main_idea}
\end{figure}

Out-of-distribution (OOD) evaluation traditionally assumes that the boundary between known and unknown inputs can be defined in advance. In classic image classification, a model is trained on a fixed set of in-distribution (ID) classes, and inputs from classes outside this set are treated as OOD \citep{hendrycks2017baseline,yang2022openood}, as illustrated in \figpanelref{fig:main_idea}{a}. This setting fits closed-world classifiers whose recognition scope is tied to a predefined training taxonomy. Vision--language pretraining changes this paradigm (\figpanelref{fig:main_idea}{b}). Vision--language models (VLMs) now learn from large collections of image--text pairs \citep{radford2021learning,jia2021scaling}, allowing them to recognize concepts beyond the labels of a single downstream task. As a result, a predefined label set no longer fully describes what a pretrained VLM can recognize. This motivates evaluating target concepts beyond a fixed downstream label set and comparing how well different models recognize them.

Existing benchmarks have revealed failures of vision--language models in specific capabilities and challenging examples, including compositional and natural-image settings \citep{hsieh2023sugarcrepe,li2024naturalbench}. However, these benchmarks are typically organized around a particular capability or failure mode, so they fail to show how recognition failures are distributed across open-world concepts. Recent OODBench studies OOD behavior in vision--language models, but it mainly considers covariate shift under a fixed ID label space, where the categories are fixed while the input distribution changes \citep{lin2026oodbench}. This leaves open which open-world concepts current models can recognize and how recognition differs across models.

To address this gap, we construct OODConceptBench, a hierarchical benchmark with two layers, to evaluate how well VLMs can recognize open-world concepts (\figpanelref{fig:main_idea}{c}). To this end, we first define a vocabulary of 560 concept phrases across seven categories. To collect images for these concepts, we use web search and image generation. After initial quality checks, we preserve 47,747 candidate images. To preserve challenging open-world cases, we then construct the benchmark using three filtering rules. The first two rules remove candidates close to common-category and rare reference banks to form the full set L1 with 30,752 images, while the third selects from L1 image--concept pairs with lower image--text similarity to the assigned concept to form the subset L2 with 14,285 images.

\begin{wrapfigure}{r}{0.50\textwidth}
\centering
\includegraphics[width=\linewidth]{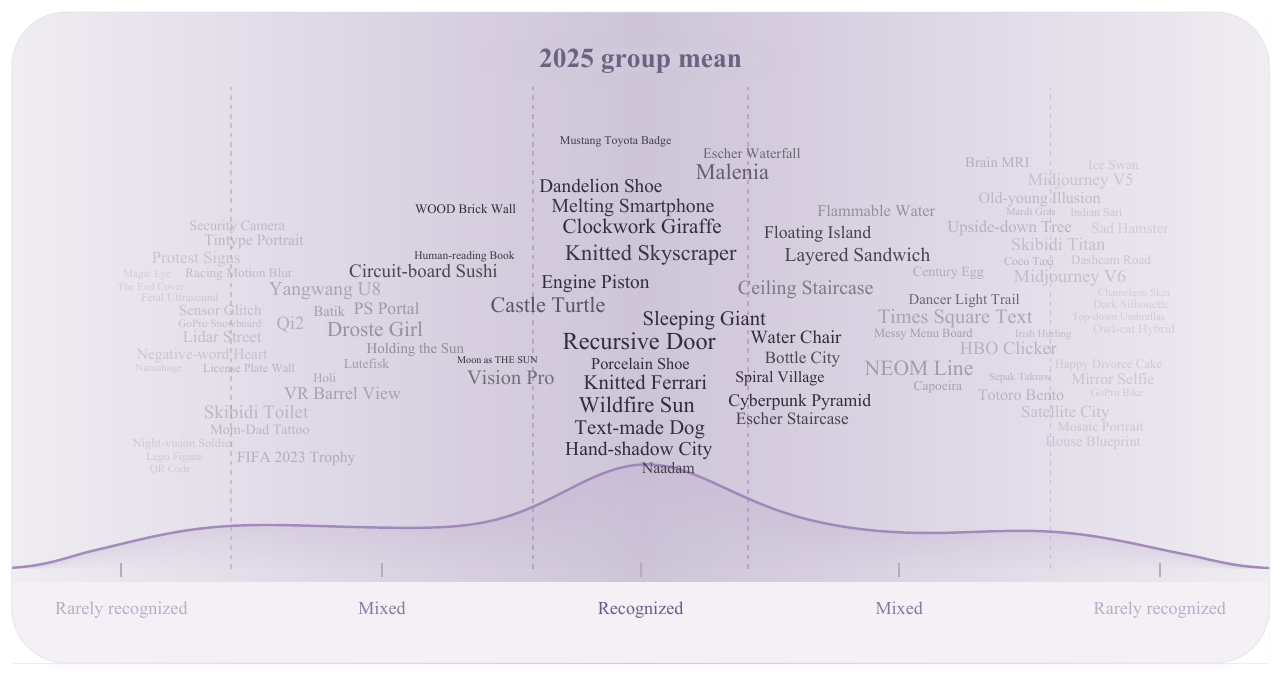}
\caption{L2 concept-recognition landscape for the 2025 release-year group. The 91 concepts are positioned by group-mean recognition, with lower recognition toward the outer regions and higher recognition toward the center.}
\label{fig:l2_wordcloud_2021_2025}
\end{wrapfigure}

We evaluate 37 VLMs across different release-year groups, model families, and variants. Throughout these evaluations, three findings stand out. First, even models in the latest release-year group show weak performance on some open-world concepts (\figref{fig:l2_wordcloud_2021_2025}), while all 37 evaluated models perform worse on L2 than on L1. Second, model performance does not improve monotonically by release year, and some later release-year groups perform worse than earlier ones. Third, an increase in overall performance does not imply consistent improvement across all benchmark categories. From 2024 to 2025, overall performance on L2 improves slightly, while five of the seven categories decline on both L1 and L2. Together, these results show that OOD remains a challenge even in the era of large-scale vision--language pretraining and motivate further study of OOD performance in VLMs.

This paper makes three main contributions.

\begingroup
\setlength{\leftmargini}{1.6em}
\setlength{\labelsep}{0.45em}
\begin{itemize}
    \item To the best of our knowledge, we are the first to systematically investigate OOD concepts for vision--language models.

    \item We introduce OODConceptBench, a hierarchical benchmark for open-world concept recognition on image--concept pairs, which covers 560 concept phrases across seven categories.

    \item We systematically evaluate 37 vision--language models and variants across release years, model families, model variants, and benchmark categories, revealing widespread OOD recognition difficulties in pretrained VLMs and underscoring the need for continued research in this area.
\end{itemize}
\endgroup

\pagebreak[4]
\section{Related Work}

\textbf{OOD evaluation under predefined boundaries.} Out-of-distribution (OOD) evaluation commonly specifies the in-distribution (ID) and OOD data before evaluation and compares methods under this fixed setting \citep{hendrycks2017baseline}. OpenOOD brings multiple OOD settings and detection methods into a unified benchmark for standardized comparison \citep{yang2022openood}. Related work further studies how models should be evaluated when known and unknown classes are considered together. OpenAUC jointly evaluates closed-set classification and open-set discrimination, while OpenworldAUC develops a unified evaluation for open-world prompt tuning that jointly considers domain detection and classification \citep{wang2022openauc,hua2025openworldauc}.

Pretrained vision--language models have also been used to improve OOD detection under specified evaluation settings. MCM uses image--text matching scores for zero-shot OOD detection, NegLabel introduces negative text labels to improve the detection signal, and InterNeg studies negative-text selection by improving consistency with the inter-modal distance used by vision--language models \citep{ming2022mcm,jiang2024negative,xu2026interneg}. These methods improve how OOD samples are detected under a given evaluation setting, but they do not directly compare how different pretrained models recognize the same concepts.

\textbf{Open-world benchmark for vision--language models.} Recent benchmarks examine vision--language models from several related perspectives. OODBench focuses on covariate-shift OOD, where the training and OOD data share the same label space but differ in input distribution. It evaluates 40K OOD instance--category pairs through a Basic-to-Advanced progression of prompted questions covering existence, counting, and logical reasoning \citep{lin2026oodbench}. Other existing benchmarks focus on specific capabilities or difficult examples: SugarCrepe studies compositional understanding, while NaturalBench collects natural examples that are difficult for strong vision--language models \citep{hsieh2023sugarcrepe,li2024naturalbench}.

These benchmarks characterize distribution shifts, specific capabilities, or difficult cases. OODConceptBench instead focuses on how recognition of the same open-world concepts varies across pretrained VLMs, using image--concept pairs for direct cross-model comparison.

\section{Problem Definition}

In predefined-class OOD evaluation, the in-distribution label set $Y_{\mathrm{ID}}$ is fixed in advance, and samples outside this predefined label space are treated as OOD. For pretrained vision--language models, however, label-set membership alone no longer fully describes what a model can recognize, because zero-shot prompting allows models to recognize concepts beyond a predefined taxonomy. Thus, $a\notin Y_{\mathrm{ID}}$ does not necessarily mean that a model $m$ cannot recognize the concept $a$. We therefore evaluate recognition directly on fixed image--concept pairs. Specifically, for a paired image $x$ and concept $a$, if the similarity between the image and the concept's text prompt does not exceed a model-specific recognition threshold, we regard the concept as OOD for model $m$ . This motivates the following model-specific recognition boundary.

\begin{definitioncard}{Model-specific recognition boundary.}
For an image--concept pair $(x_i,a_i)$ and a vision--language model $m$, let $s_m(x_i,a_i)$ denote the similarity between image $x_i$ and the text prompt for concept $a_i$, and let $\tau_m$ denote the recognition threshold for model $m$. We define
\[
R_m(x_i,a_i)=
\begin{cases}
1, & s_m(x_i,a_i)>\tau_m,\\
0, & s_m(x_i,a_i)\leq\tau_m.
\end{cases}
\]
We count the concept as recognized when $R_m(x_i,a_i)=1$. When $R_m(x_i,a_i)=0$, the pair lies on the OOD side of model $m$'s recognition boundary. In our evaluation, $s_m(x_i,a_i)$ is instantiated as the image--concept cosine similarity and $\tau_m$ as the model-specific threshold $Q_{95,m}$; the calibration procedure is described in Section~6.1, with its finite-sample guarantee provided in \appref{app:evaluation-theory}. Because this boundary is model-specific, the same image--concept pair may be recognized by one model but not another.
\end{definitioncard}

\section{Benchmark Construction}

Building on the OOD definition above, we next describe the construction of OODConceptBench.

\subsection{Target Vocabulary, Image Collection, and Initial Checks}

Because our evaluation operates on fixed image--concept pairs, benchmark construction begins by defining the concept vocabulary and then collecting candidate images for each concept.

Following the prompt template used by CLIP \citep{radford2021learning}, each vocabulary item is expressed as a short concept phrase. We organize the vocabulary into seven diagnostic categories: \catname{Time-lagged}, \catname{Synthetic / fictional}, \catname{Cultural blindspots}, \catname{Fine-grained / texture}, \catname{Environmental shift}, \catname{Viewpoint / scale}, and \catname{Typographic conflict}. Notably, \catname{Time-lagged} refers to concepts affected by temporal change, while \catname{Typographic conflict} refers to cases where textual and visual cues conflict. Each category contains eight concept sets, with ten concept phrases per set, resulting in 80 phrases per category and 560 phrases in total. This design keeps the vocabulary balanced across the seven categories.

We collect candidate images from two sources: web search and image generation. Web search is used when suitable real-world examples can be retrieved reliably, while image generation supplements concepts for which such examples are difficult to obtain. We preserve the acquisition source of each image for subsequent source-aware analysis, and concept sets containing images from both sources are recorded as hybrid. \figpanelref{fig:dataset-distributions}{b} summarizes the resulting source composition, and \appref{app:source-aware-diagnostics} reports additional source-based analyses.

Before applying the filtering rules, we manually review the collected images for technical usability and remove corrupted, unreadable, or otherwise unusable files. This review preserves the concept assignments established during data collection. Ordinary retrieval noise from web search is also retained at this stage and is handled by the subsequent filtering procedure. The remaining 47,747 images form L0, the initial candidate pool for benchmark construction. A subsequent format check excludes 914 files, leaving 46,833 candidates to enter the three filtering rules.

\subsection{Filtering Rules}

We then apply three filtering rules to these candidates. All three rules compare candidate images against specific reference banks using image-text similarity, and we use OpenAI CLIP ViT-B/32 \citep{radford2021learning} for all three filtering rules.

\begin{tcolorbox}[filteringrulescard]
\noindent\textbf{Filtering rules.} The three filtering rules serve different roles:\par\vspace{0.8mm}
{\color{FilteringRuleHeading}\bfseries Filtering rule 1 --- Common-category reference.} Candidates close to the common-category reference bank are removed.\par\vspace{0.7mm}
{\color{FilteringRuleHeading}\bfseries Filtering rule 2 --- Rare reference.} The remaining candidates are further filtered using the Rare reference bank, and the retained images form L1.\par\vspace{0.7mm}
{\color{FilteringRuleHeading}\bfseries Filtering rule 3 --- Target-concept alignment.} Using the construction model, we select from L1 the images with lower similarity to their assigned concepts to form the L2 subset.
\end{tcolorbox}

Let \(z(x)\) denote the normalized image representation, \(B_{1K}\) the fixed ImageNet-1K common-category reference bank, \(B_{21K}\) the fixed 200-vector Rare reference bank derived from an ImageNet-21K semantic source \citep{deng2009imagenet}, and \(t(a)\) the normalized text representation of concept \(a\). We define
\[
\begin{aligned}
S_{1K}(x)&=1-\max_{b\in B_{1K}} z(x)^\top b,\\
S_{21K}(x)&=1-\max_{b\in B_{21K}} z(x)^\top b,\\
S_{\mathrm{Txt}}(x,a)&=z(x)^\top t(a).
\end{aligned}
\]
Larger \(S_{1K}\) and \(S_{21K}\) indicate greater distance from the corresponding reference bank, while lower \(S_{\mathrm{Txt}}\) indicates lower similarity to the assigned concept. Filtering rule 1 applies \(S_{1K}\), filtering rule 2 applies \(S_{21K}\) to the survivors of rule 1, and filtering rule 3 applies \(S_{\mathrm{Txt}}\) to the survivors of rule 2. The thresholds for the three criteria are determined by the calibration procedures described in \appref{app:threshold-calibration}. \tabref{tab:gate-definitions} summarizes the filtering criteria and the resulting benchmark sizes.

\begin{table}[!htbp]
\caption{Filtering criteria and resulting benchmark sizes.}
\label{tab:gate-definitions}
\centering
\setlength{\MainTableHeaderWidth}{\linewidth}
\begin{tabularx}{\linewidth}{@{}X>{\centering\arraybackslash}p{0.27\linewidth}>{\centering\arraybackslash}p{0.18\linewidth}@{}}
\MainTableHeaderBackground\textbf{Filtering rule} & \textbf{Criterion} & \textbf{Result} \\[2pt]
Rule 1 --- Common-category reference & $S_{1K}>0.7175$ & 34,963 \\
Rule 2 --- Rare reference & $S_{21K}>0.2976$ & L1: 30,752 \\
Rule 3 --- Target-concept alignment & $S_{\mathrm{Txt}}<0.3031$ & L2: 14,285 \\
\bottomrule
\end{tabularx}
\end{table}

\FloatBarrier

 A WordNet expansion check further supports broad reference-vocabulary coverage; details are provided in \appref{app:wordnet-expansion-check}.

\FloatBarrier

\section{Benchmark Composition and Coverage}

\begin{figure}[!h]
\centering
\includegraphics[width=\linewidth]{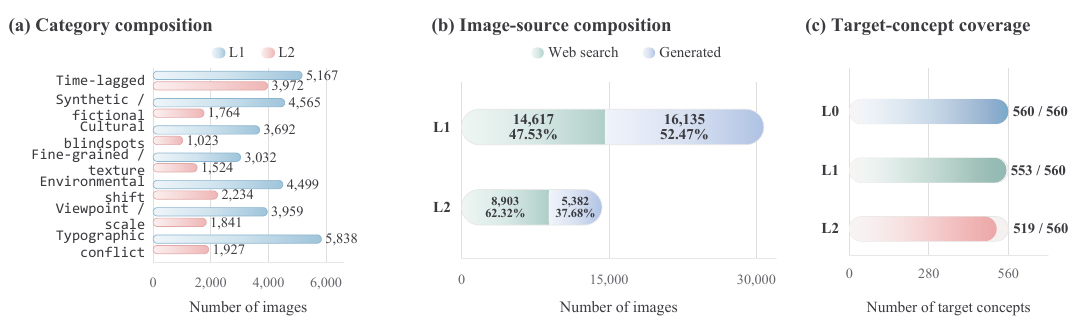}
\caption{Benchmark composition and coverage. (a) Category counts in L1 and L2, (b) image-source composition, and (c) concept coverage, measured by the number of target concepts with at least one image in L0, L1, and L2.}
\label{fig:dataset-distributions}
\end{figure}

Both L1 and L2 contain all seven categories but differ in category composition, image-source composition, and concept coverage. \figref{fig:dataset-distributions} summarizes these differences.

All seven categories are represented in both L1 and L2, although their sizes differ. In L1, category counts range from 3,032 for \catname{Fine-grained / texture} to 5,838 for \catname{Typographic conflict}. In L2, they range from 1,023 for \catname{Cultural blindspots} to 3,972 for \catname{Time-lagged}. Representative L1-only and L2 image--concept pairs from all seven categories are shown in \appfigref{fig:F1-real-examples}.

L1 and L2 also differ in image-source composition and concept coverage. Web-search images account for 47.53\% of L1 and 62.32\% of L2. Of the 560 concept phrases, L0 covers all 560, while L1 covers 553 and L2 covers 519. Additional category-retention results and source-based analyses are provided in \appref{app:category-retention} and~\ref{app:source-aware-diagnostics}.

\section{Model Evaluation}

After construction, L1/L2 membership is fixed before model evaluation. We evaluate 37 vision--language models spanning release-year groups, model families, and variants on the same fixed L1 and L2 benchmark sets. Each model is evaluated with its own image–text representation and an independently calibrated model-specific threshold \(Q_{95,m}\). The Filtering rule 3 score is used only to determine L2 membership.

\subsection{Evaluation Protocol}

Because the evaluated models differ in architecture and other design choices, their raw image--text similarity scores may lie on different numerical scales. We therefore use the same reference image pool $C$ and text bank $P$ to calibrate a model-specific recognition threshold for each model.

The reference image pool \(C\) contains \(n=2{,}000\) images, with 500 images each from MNIST, CIFAR-100, STL-10, and Oxford-IIIT Pet \citep{lecun1998mnist,krizhevsky2009cifar,coates2011stl,parkhi2012cats}. Using the same number of images from each dataset gives the four sources equal weight and provides every model with the same diverse reference pool for calibration. The text bank \(P\) contains the 553 target concept phrases with at least one image in L1. All phrases use the same template, ``a photo of \textless concept\textgreater'', which keeps the textual context fixed across models and avoids concept- or model-specific prompt tuning.

For each calibration image \(X_i\), we first measure how strongly it matches the target vocabulary by taking its maximum similarity over the prompt bank \(P\):
\[
Z_{m,i}
=
\max_{p\in P}
\cos\!\left(f_m^I(X_i),f_m^T(p)\right).
\]

where \(f_m^I\) and \(f_m^T\) denote the image and text encoders of model \(m\), respectively. Taking the maximum accounts for the possibility that a background image may obtain high similarity to any target prompt rather than only to a particular one. 

We then use these calibration scores to determine a model-specific recognition threshold. With \(n=2{,}000\) calibration images and \(\alpha=0.05\), we set
\[
Q_{95,m}=Z_{m,(k)},
\qquad
k=\left\lceil(n+1)(1-\alpha)\right\rceil=1901,
\]
where \(Z_{m,(k)}\) is the \(k\)-th smallest calibration score. This is the standard finite-sample quantile correction used in conformal inference \citep{DBLP:journals/ftml/AngelopoulosB23}. Under the exchangeability condition stated in \appref{app:finite-sample-control}, this finite-sample calibration controls the probability that a new background image exceeds the threshold for at least one target prompt at level \(\alpha\).

With \(Q_{95,m}\) fixed for each model, we then evaluate the benchmark image--concept pairs. For a benchmark image \(x_i\) with assigned target concept \(a_i\) and corresponding prompt \(p_i\), Definition 1 is instantiated as
\[
R_m(x_i,a_i)
=
\mathbf{1}
\left[
\cos\!\left(f_m^I(x_i),f_m^T(p_i)\right)
>
Q_{95,m}
\right],
\]
where \(\mathbf{1}[\cdot]\) is the indicator function. Thus, the assigned concept is counted as recognized when its image--prompt similarity exceeds the independently calibrated threshold for model \(m\).

We use the same calibration image pool, target prompt bank, value of \(\alpha\), and calibration rule for all models, while allowing the resulting threshold \(Q_{95,m}\) to vary by model. The recognition rate is then computed as the proportion of successfully processed benchmark samples for which \(R_m(x_i,a_i)=1\). Processing failures are recorded separately and excluded from the denominator. \appref{app:evaluation-theory} provides the finite-sample guarantee and the corresponding invariance property, while \appref{app:full-cross-model-results} reports the model-specific thresholds and full evaluation results.

\subsection{L2 Recognition Remains Lower Than L1 Across the Evaluated Models}

\rqbox{RQ1}{How well can current vision--language models recognize the target concepts in the benchmark?}

Filtering rule 3 establishes the L1/L2 ordering using the fixed construction model. The cross-model evaluation then tests whether this ordering persists across independently evaluated models.

Under this evaluation, the overall recognition rate on L2 is lower than that on L1 for each of the 37 evaluated models. When the results are further broken down by model and benchmark category, 258 of the 259 comparisons show the same direction; the only exception is a near tie in \catname{Time-lagged}. This shows that lower recognition on L2 is not confined to a small number of models or benchmark categories.

\begin{figure*}[!htbp]
\centering
\includegraphics[width=\textwidth]{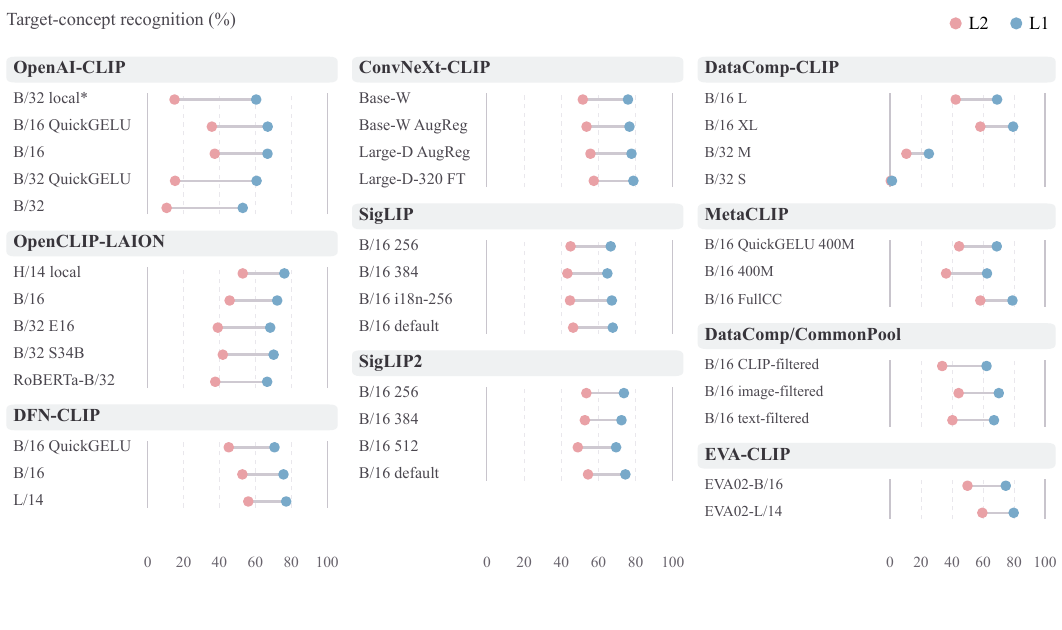}
\caption{Per-model target-concept recognition on L1 and L2. Models are grouped by family, and each row compares recognition on L2 and L1 for the same model. Across all 37 evaluated models, recognition is lower on L2 than on L1. * B/32 local denotes the OpenAI CLIP ViT-B/32 checkpoint used as the fixed construction model.}
\label{fig:section6_model_matrix}
\end{figure*}

At the same time, recognition rates vary widely across the evaluated models, ranging from 1.3\% to 79.8\% on L1 and from 0.5\% to 59.6\% on L2. No evaluated model reaches 80\% recognition on L1 or 60\% on L2. \figref{fig:section6_model_matrix}  reports the per-model target-concept recognition results on L1 and L2, grouped by model family.

To further examine the L1/L2 difference observed in the recognition results, we perform two complementary diagnostics using the fixed construction model. For the fixed construction model, the mean entropy of the prediction distributions increases from 3.724 on L1 to 3.757 on L2, indicating greater prediction uncertainty on L2. Target-semantic loss, defined as one minus the cosine similarity between each image and its assigned target concept, also increases from 0.697 to 0.736, indicating weaker target-concept alignment on L2. Both diagnostics therefore show the same L1/L2 direction as the recognition results and provide additional evidence from the fixed construction model. These two diagnostics describe the fixed construction model, while the cross-model evidence for Finding 1 comes from the model- and category-level recognition results above.

\findingbox{1}{L2 recognition remains lower than L1 across the evaluated models.}{After L1/L2 membership is fixed, recognition is lower on L2 than on L1 for all 37 evaluated models, and 258 of the 259 model--benchmark-category comparisons show the same direction. Thus, the L1--L2 ordering established during construction persists across the evaluated models under the common evaluation protocol.}

\subsection{Later-Released Models Do Not Necessarily Show Stronger Open-World Target-Concept Recognition}

\rqbox{RQ2}{How do target-concept recognition rates differ across release-year groups, model families, and variants?}

\begin{figure*}[t]
\centering
\includegraphics[width=\textwidth]{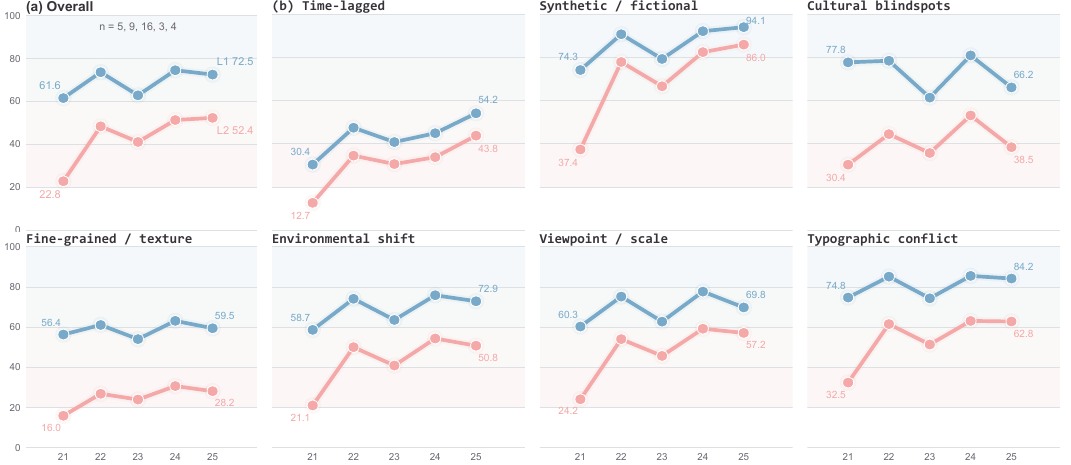}
\caption{Recognition across release-year groups and benchmark categories. (a) Mean L1 and L2 recognition rates for models in each release-year group from 2021 to 2025. (b) The same comparison for each of the seven benchmark categories. All panels use the same 0--100 scale and recognition-range background bands.}
\label{fig:section6_year_category_matrix}
\end{figure*}

Recognition does not increase steadily with release year. On L2, the average recognition rates for the 2021--2025 release-year groups are 22.8\%, 48.4\%, 41.1\%, 51.4\%, and 52.4\%, respectively. On L1, the corresponding rates are 61.6\%, 73.6\%, 62.9\%, 74.6\%, and 72.5\%. Both layers therefore decline from 2022 to 2023, while L1 also declines from 2024 to 2025. \figpanelref{fig:section6_year_category_matrix}{a} shows these release-year trends.

Differences also remain within model families. On L2, the selected ConvNeXt variants differ by about 4.0 percentage points, the EVA variants by 9.6 points, and the OpenAI CLIP variants by 23.0 points. Some later variants have higher recognition. For example, the SigLIP2 mean is 52.4\%, compared with 44.9\% for SigLIP. However, later variants do not consistently have higher recognition across the evaluated comparisons.

Models released in the same year can also differ in recognition. Among the 2024 DFN variants, L2 recognition is 52.8\% for B/16 standard, 45.2\% for B/16 QuickGELU, and 56.1\% for L/14. Release year alone therefore does not order model performance.

\findingbox{2}{Later-released models do not necessarily achieve stronger open-world target-concept recognition.}{These results suggest that stronger open-world target-concept recognition may require models and training algorithms specifically designed for OOD recognition, rather than relying on general model updates alone.}

\subsection{An Increase in Overall Recognition Does Not Imply Consistent Improvement Across All Benchmark Categories}

\rqbox{RQ3}{When overall recognition increases, does recognition also increase in every benchmark category on L1 and L2?}

Recognition rates differ across the seven benchmark categories. Averaged across all 37 evaluated models, L2 recognition ranges from 24.6\% for \catname{Fine-grained / texture} to 68.9\% for \catname{Synthetic / fictional}. The L1--L2 difference also varies by category, from 12.0 percentage points for \catname{Time-lagged} to 32.8 points for \catname{Fine-grained / texture}; \catname{Cultural blindspots} shows a similarly large difference of 31.0 points. A single overall recognition rate therefore cannot capture the differences across the seven benchmark categories.

From 2024 to 2025, the overall L2 recognition rate increases slightly from 51.4\% to 52.4\%. However, recognition increases on both L1 and L2 only for \catname{Time-lagged} and \catname{Synthetic / fictional}, while it decreases for the other five benchmark categories. \figpanelref{fig:section6_year_category_matrix}{b} shows these category-level changes.

Comparing the 2021 and 2025 release-year groups, recognition on L2 is higher in 2025 for all seven benchmark categories, with increases ranging from 8.1 to 48.6 percentage points. On L1, recognition is higher in six categories, while \catname{Cultural blindspots} is 11.6 points lower in 2025. Thus, the category-level changes from 2021 to 2025 are not identical on L1 and L2.

\findingbox{3}{An increase in overall recognition does not imply consistent improvement across all benchmark categories.}{Improvements in overall recognition can coexist with declines in individual benchmark categories. Therefore, category-level results are necessary to reveal performance differences that may be obscured by a single aggregate score. }

\subsection{Supporting Analyses}

We conduct three supporting analyses to examine whether the L1/L2 recognition difference can be explained by simpler alternatives. Specifically, we consider conventional fixed-ID OOD behavior, semantic distance from common visual categories, and learnability under limited supervision. Detailed protocols and results are provided in \appref{app:supporting-analyses}.

\textbf{Fixed-ID comparison.} We ask whether the L1/L2 difference observed under our target-concept recognition protocol can also be captured by conventional fixed-ID OOD evaluation. Using Oxford-IIIT Pet as the fixed ID reference, mean confidence is 0.547 on ID images and 0.263/0.261 on L1/L2. AUROC is 0.830/0.831 and FPR95 is 0.650/0.658, with similarly small L1/L2 differences for CIFAR-100 and STL-10. These results show that fixed-ID evaluation separates the benchmark samples from reference ID data, but does not capture the L1/L2 concept recognition difference observed in Section~6.2.

\textbf{Semantic relation to ImageNet-1K.} A possible concern is that L2 is more difficult simply because its target concepts are more semantically distant from common visual categories. We therefore measure each target concept's similarity to its nearest ImageNet-1K class in the fixed CLIP ViT-B/32 text space. The median similarity is 0.709 for L1 and 0.710 for L2. The nearly identical values suggest that the lower recognition observed on L2 cannot be explained simply by greater semantic distance from the ImageNet-1K taxonomy.

\textbf{Limited-supervision classification.} A possible concern is that the low target-concept recognition observed on the benchmark may simply reflect that its samples are not learnably distinguishable from ordinary control images in the feature space. To examine this, we train balanced binary logistic-regression classifiers on frozen CLIP features using 1--30 labeled examples per class. Mean accuracy increases from 0.627 to 0.887 on L1 and from 0.603 to 0.867 on L2. These results show that the benchmark samples are distinguishable from separately collected control images under limited supervision.

Together, these analyses suggest that the target-concept recognition difficulty observed on L1 and L2 is not simply explained by conventional fixed-ID OOD behavior, greater semantic distance from ImageNet-1K, or a lack of learnable structure in the benchmark samples.

\section{Conclusion}

We introduce OODConceptBench, a hierarchical benchmark for open-world target-concept recognition on fixed image–concept pairs. Instead of defining OOD solely by membership in a predefined label set, OODConceptBench evaluates whether each model can recognize an assigned concept under a model-specific recognition boundary. Across 37 vision–language models, target-concept recognition is consistently lower on L2 than on L1. Performance also does not improve monotonically across release-year groups, and gains in overall recognition can coincide with declines in individual benchmark categories. Supporting analyses further suggest that the target-concept recognition difficulty observed on L1 and L2 is not simply explained by conventional fixed-ID OOD behavior, greater semantic distance from ImageNet-1K, or poor learnability of the benchmark samples. Taken together, OODConceptBench provides a common basis for diagnosing open-world recognition behavior across pretrained VLMs and motivates the development of models and training algorithms specifically designed for OOD recognition.

\section*{LLM Usage}
Generative AI tools, including large language models, were used to assist with language editing; drafting and revising parts of the manuscript; document organization; literature retrieval and reference checking; feedback on experimental design and interpretation of results; and code generation and review for data processing, evaluation, and plotting. Image-generation tools were also used to create part of the generated-image component of the benchmark, as described in Section~4.1. The authors made all final methodological and experimental decisions and independently reviewed and verified the data processing, numerical results, figures, citations, and scientific claims. Generative AI tools were not used to prove mathematical claims. The authors take responsibility for the final content of the paper and all AI-assisted research artifacts.

\section*{Reproducibility Statement}
Benchmark construction is described in Section~4, with calibration details provided in Appendix~A. The evaluation protocol is defined in Section~6.1, with its finite-sample guarantee provided in Appendix~B. Full model- and category-level results are reported in Appendix~C, supporting analyses in Appendix~D, and the release boundary and scope are summarized in Appendix~E.

\section*{Ethics Statement}
This benchmark combines web-search and generated images for research on open-world concept recognition. As detailed in Appendix~\ref{app:reproducibility-release-boundary}, any image-level public release is conditioned on near-duplicate, sensitive-content, licensing, source-policy, and privacy review. Generated images will remain clearly marked, source information will be retained for web-search images, and local paths and other environment-specific information will be removed from released research files.

\bibliographystyle{05_paper/arxiv_references}
\bibliography{05_paper/references}

\clearpage
\appendix
\renewcommand{\thetable}{\Alph{section}\arabic{table}}
\renewcommand{\thefigure}{\Alph{section}\arabic{figure}}
\renewcommand{\theHtable}{\Alph{section}.\arabic{table}}
\renewcommand{\theHfigure}{\Alph{section}.\arabic{figure}}
% Historical appendix visual language restored from version 50.
% The current A--E structure and single-column navigation are retained.
\definecolor{HistoricalConvNeXtBand}{HTML}{E3DDD4}
\definecolor{HistoricalDFNBand}{HTML}{D9E1D7}
\definecolor{HistoricalDataCompBand}{HTML}{E4DADD}
\definecolor{HistoricalEVABand}{HTML}{DED9E6}
\definecolor{HistoricalMetaBand}{HTML}{D8E2E0}
\definecolor{HistoricalOpenAIBand}{HTML}{E5E1D2}
\definecolor{HistoricalCommonPoolBand}{HTML}{DFE3D5}
\definecolor{HistoricalOpenCLIPBand}{HTML}{D8DDE6}
\definecolor{HistoricalSigLIPBand}{HTML}{E5DDD5}
\definecolor{HistoricalSigLIP2Band}{HTML}{D5E2DE}

% Ordinary appendix tables remain black-and-white.
\newcommand{\AppendixTableHead}[1]{{\scriptsize\textbf{#1}}}
\newcommand{\AppendixFamilyTag}[1]{\textbf{#1}}

% Only the two full model-category tables use the historical family bands.
\newcommand{\AppendixFamilyBandRow}[1]{%
  \begingroup
  \def\AppendixFamilyBandColor{HistoricalConvNeXtBand}%
  \ifstrequal{#1}{ConvNeXt-CLIP}{\def\AppendixFamilyBandColor{HistoricalConvNeXtBand}}{}%
  \ifstrequal{#1}{DFN-CLIP}{\def\AppendixFamilyBandColor{HistoricalDFNBand}}{}%
  \ifstrequal{#1}{DataComp-CLIP}{\def\AppendixFamilyBandColor{HistoricalDataCompBand}}{}%
  \ifstrequal{#1}{EVA-CLIP}{\def\AppendixFamilyBandColor{HistoricalEVABand}}{}%
  \ifstrequal{#1}{MetaCLIP}{\def\AppendixFamilyBandColor{HistoricalMetaBand}}{}%
  \ifstrequal{#1}{OpenAI-CLIP}{\def\AppendixFamilyBandColor{HistoricalOpenAIBand}}{}%
  \ifstrequal{#1}{DataComp/CommonPool}{\def\AppendixFamilyBandColor{HistoricalCommonPoolBand}}{}%
  \ifstrequal{#1}{OpenCLIP-LAION}{\def\AppendixFamilyBandColor{HistoricalOpenCLIPBand}}{}%
  \ifstrequal{#1}{SigLIP}{\def\AppendixFamilyBandColor{HistoricalSigLIPBand}}{}%
  \ifstrequal{#1}{SigLIP2}{\def\AppendixFamilyBandColor{HistoricalSigLIP2Band}}{}%
  \colorbox{\AppendixFamilyBandColor}{%
    \parbox{\dimexpr\linewidth-2\fboxsep\relax}{\strut\textbf{#1}}}%
  \endgroup
}

% The historical Roadmap used the existing rounded roadmapgroup style and
% purple-gray paper palette; every current A--E entry now shares that language.
\newcommand{\AppendixRoadmapEntry}[3]{%
  \begin{tcolorbox}[roadmapgroup,colback=RoadmapSupporting]
  \noindent{\color{PaperLinkPurple}\bfseries Appendix #1} ---
  {\color{RoadmapHeading}\bfseries #2}\par\smallskip
  {\color{RoadmapText}#3}
  \end{tcolorbox}%
}

\clearpage
\section*{Appendix Roadmap}
\vspace{0.4em}
\begingroup
\small
\AppendixRoadmapEntry{A}{Benchmark Construction Details}{Calibration settings, threshold stability, category-wise retention, and qualitative examples.}
\AppendixRoadmapEntry{B}{Evaluation Theory}{Finite-sample calibration guarantee and invariance to strictly increasing score transformations.}
\AppendixRoadmapEntry{C}{Full Cross-model Results}{Model registry, complete L1/L2 category-level results, release-year and variant analyses, and recognition maps.}
\AppendixRoadmapEntry{D}{Supporting Analyses}{Fixed-ID comparison, limited-supervision classification, source diagnostics, threshold sensitivity, fixed-model diagnostics, and ImageNet-1K semantic analyses.}
\AppendixRoadmapEntry{E}{Release Boundary and Scope}{Intended use, release conditions, and study limitations.}
\endgroup
\clearpage

\section{Benchmark Construction Details}
\label{app:construction}
\label{app:benchmark-construction-details}
\label{app:threshold-calibration}
\setcounter{table}{0}
\setcounter{figure}{0}

\subsection{Threshold Calibration Settings}
\label{app:construction-calibration-settings}

We use the construction scores defined in Section~4.2. Table~\ref{tab:app-threshold-calibration} summarizes the calibration data, statistics, rules, and fixed thresholds.

\begin{table}[!htbp]
\centering
\footnotesize
\caption{Calibration settings and final thresholds for the three benchmark-construction rules.}
\label{tab:app-threshold-calibration}
\renewcommand{\arraystretch}{1.10}
\setlength{\tabcolsep}{4pt}
\begin{tabularx}{0.97\textwidth}{@{}>{\centering\arraybackslash}p{0.07\textwidth}>{\raggedright\arraybackslash}p{0.44\textwidth}>{\raggedright\arraybackslash}p{0.27\textwidth}>{\centering\arraybackslash}X@{}}
\toprule
\AppendixTableHead{\shortstack{Filtering\\rule}} & \AppendixTableHead{Calibration setting} & \AppendixTableHead{Calibration rule} & \AppendixTableHead{Final threshold} \\
\midrule
1 & Up to 3,000 ID calibration images; ImageNet-1K validation with Tiny-ImageNet validation fallback; ImageNet-1K reference features. & 95th percentile of $S_{\mathrm{1K}}$ calibration scores. & $\boldsymbol{T_1=0.7175}$ \\
2 & Filtering-rule-1 calibration survivors satisfying $S_{\mathrm{1K}}>T_1$, plus up to 1,500 Tiny-ImageNet training images; fixed 200-vector long-tail reference bank. & 95th percentile of the mixed calibration-pool $S_{\mathrm{21K}}$ scores. & $\boldsymbol{T_2=0.2976}$ \\
3 & 5,000 MNIST test images as negative samples; fixed 200-concept calibration vocabulary with repeated subsampling. & Per-round candidate threshold from the maximum target-prompt similarity plus its standard deviation, followed by the multi-round stability check. & $\boldsymbol{T_3=0.3031}$ \\
\bottomrule
\end{tabularx}
\end{table}

\FloatBarrier

\subsection{Filtering Rule 1 and Filtering Rule 2 Calibration Results}
\label{app:construction-gate12}

\begin{figure}[!htbp]
  \centering
  \includegraphics[width=0.49\linewidth]{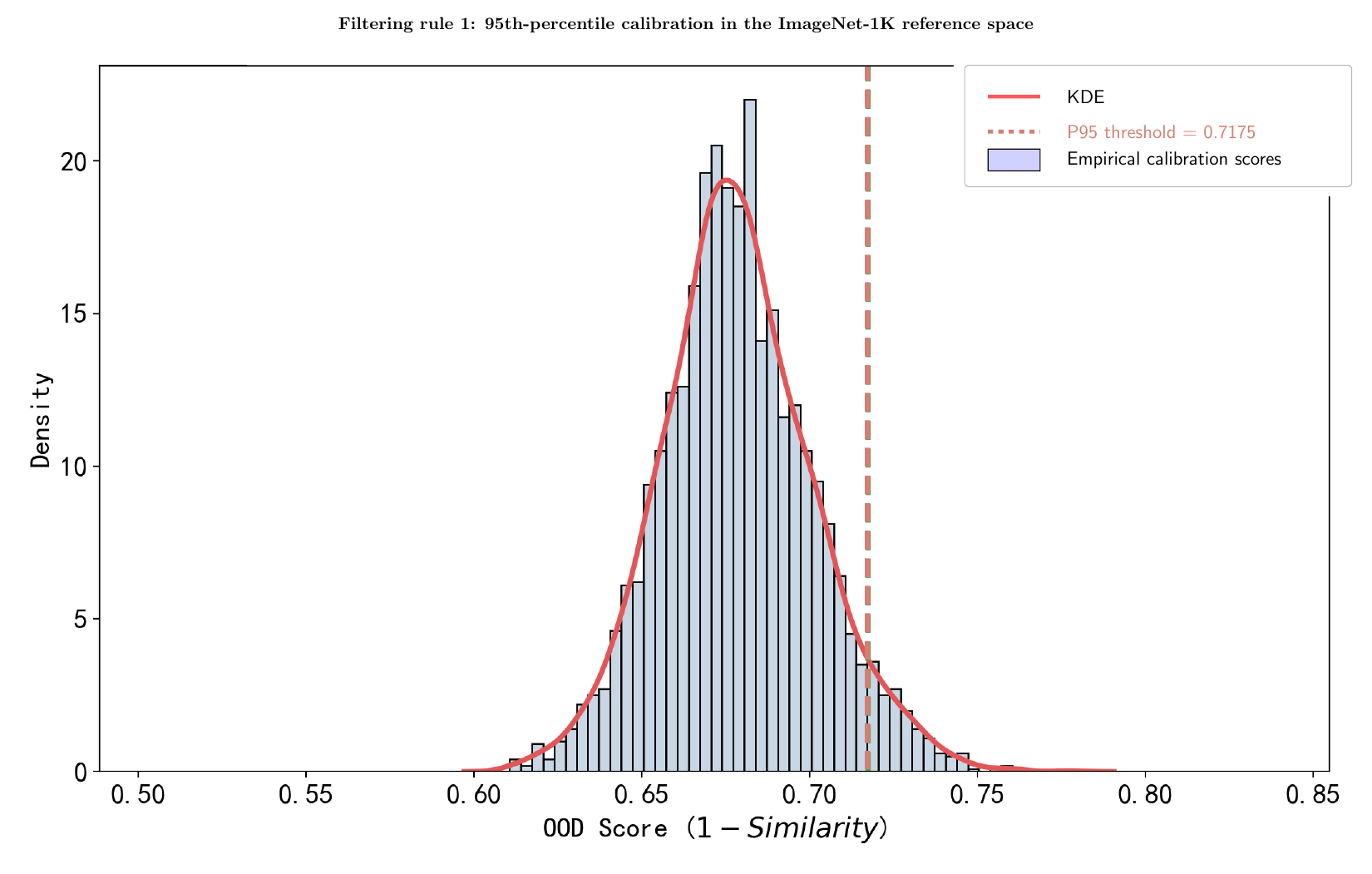}\hfill
  \includegraphics[width=0.49\linewidth]{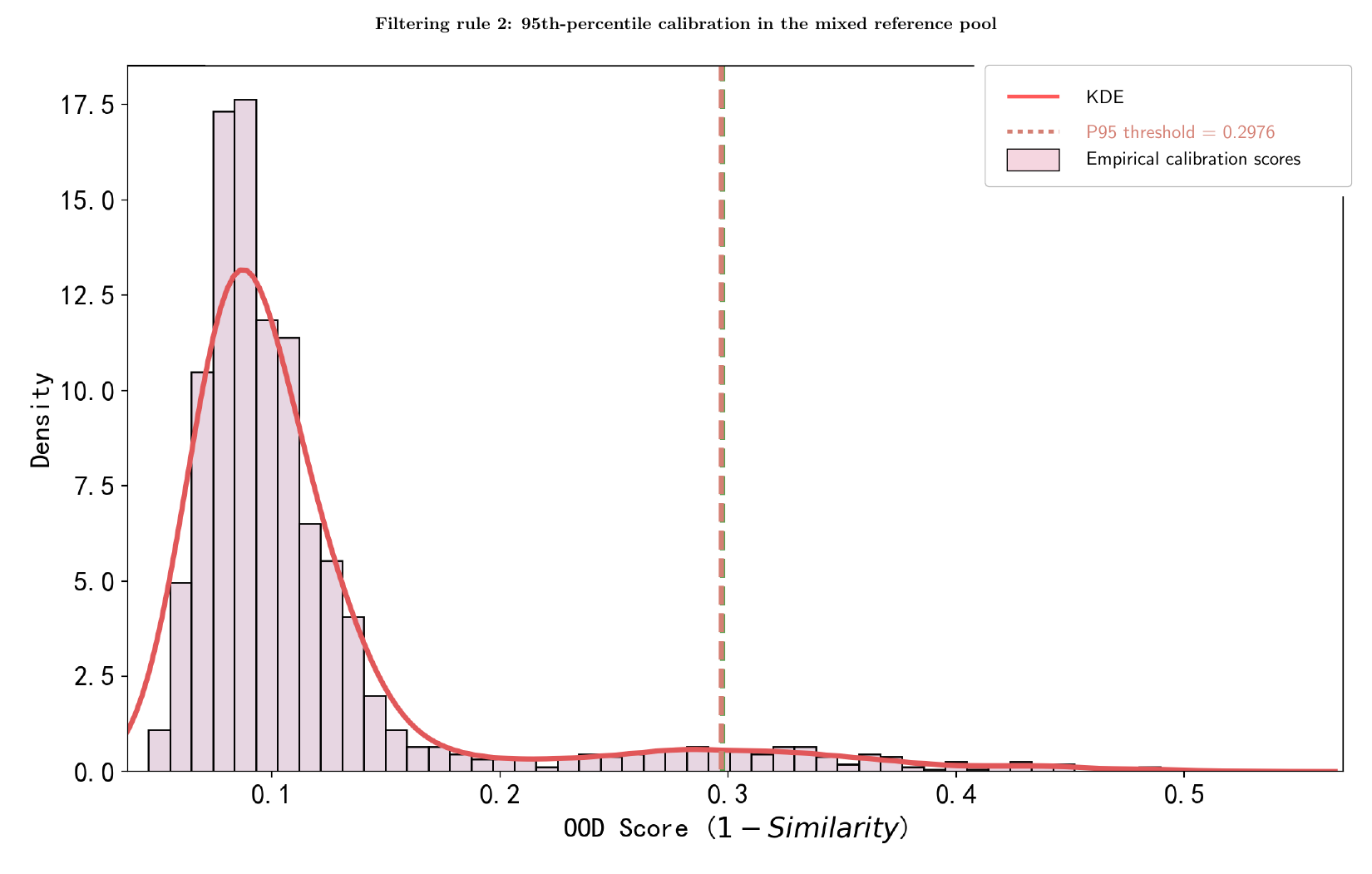}
  \caption{Filtering rule 1 and filtering rule 2 calibration-score distributions. The vertical lines mark the 95th-percentile thresholds, $T_1=0.7175$ and $T_2=0.2976$; KDE curves are visual summaries only.}
  \label{fig:app-c-calibration-distributions}
\end{figure}

Filtering rules 1 and 2 use the 95th percentile of their corresponding calibration distributions. Figure~\ref{fig:app-c-calibration-distributions} shows the empirical score distributions and threshold lines; KDE visualizes distribution shape but does not determine either threshold.

\FloatBarrier

\subsection{Filtering Rule 3 Calibration Stability}
\label{app:construction-gate3}

Filtering rule 3 uses 5,000 MNIST test images as negative samples and a calibration vocabulary of 20 groups and 200 concept phrases. The calibration was repeated over five rounds using \texttt{SAMPLE\_STEPS=\{1,2,3,5,8,10,20,50\}}. For each round and text subset, every image takes its maximum similarity $m_i$, and the candidate threshold is
\[
\max_i(m_i)+\operatorname{Std}_i(m_i).
\]
Step 10 includes all 10 phrases in each group, so steps 20 and 50 also use 200 prompts. The full-vocabulary mean is 0.303085, reported as the construction threshold $T_3=0.3031$.

\begin{figure}[!htbp]
\centering
\begin{minipage}[t]{.42\linewidth}\vspace{0pt}
\begingroup\makeatletter\def\@captype{table}\makeatother
\centering

\small
\setlength{\tabcolsep}{1.8pt}
\caption{Filtering rule 3 repeated-sampling threshold stability.}\label{tab:app-t3-stability}
\vspace{0.35em}

\renewcommand{\arraystretch}{1.08}
\begin{tabular*}{0.98\linewidth}{@{\extracolsep{\fill}}rrrr@{}}
\toprule
\AppendixTableHead{\shortstack{Phrases per\\group}} & \AppendixTableHead{\shortstack{Actual total\\prompts}} & \AppendixTableHead{Mean $T_3$} & \AppendixTableHead{Std.} \\
\midrule
1 & 20 & 0.274397 & 0.010029 \\
2 & 40 & 0.284339 & 0.006683 \\
3 & 60 & 0.301668 & 0.003141 \\
5 & 100 & 0.301752 & 0.003368 \\
8 & 160 & 0.303244 & 0.000212 \\
10 & 200 & 0.303085 & 0.000000 \\
20$^{*}$ & 200 & 0.303085 & 0.000000 \\
50$^{*}$ & 200 & 0.303085 & 0.000000 \\
\bottomrule
\end{tabular*}

\par\endgroup

\end{minipage}\hfill
\begin{minipage}[t]{.54\linewidth}\vspace{0pt}
\caption{Filtering rule 3 threshold stability across sampling settings.}
\label{fig:app-c-t3-stability}
\includegraphics[width=\linewidth]{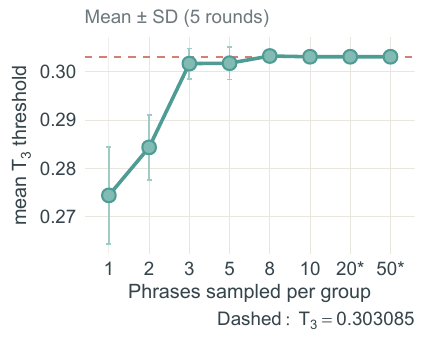}
\end{minipage}
\par\smallskip{\footnotesize $^{*}$Each group contains at most 10 phrases, so settings 20 and 50 still use 200 prompts in total.}
\end{figure}
\FloatBarrier

\subsection{Standard-dataset Behavior Check}
\label{app:construction-standard-check}

This post-hoc analysis checks the behavior of the fixed $T_1$ and $T_2$ thresholds on standard datasets. These datasets are not used to select the thresholds.

\begin{center}
\refstepcounter{table}\label{tab:app-standard-dataset-validation}
\scriptsize
\setlength{\tabcolsep}{4pt}
\textbf{Table~\thetable: Validation of the fixed filtering-rule-1 and filtering-rule-2 thresholds on standard datasets.}
\vspace{0.35em}

\renewcommand{\arraystretch}{1.10}
\begin{tabular*}{0.90\textwidth}{@{\extracolsep{\fill}}lrrrrr@{}}
\toprule
\AppendixTableHead{Dataset} & \AppendixTableHead{Total} & \AppendixTableHead{After rule 1} & \AppendixTableHead{Rule-1 filter rate} & \AppendixTableHead{After rules 1+2} & \AppendixTableHead{Total filter rate} \\
\midrule
CIFAR-10 & 10,000 & 6,021 & 39.79\% & 11 & 99.89\% \\
MNIST & 10,000 & 9,992 & 0.08\% & 0 & 100.00\% \\
CIFAR-100 & 10,000 & 6,207 & 37.93\% & 11 & 99.89\% \\
STL-10 & 8,000 & 3,582 & 55.23\% & 326 & 95.93\% \\
SVHN & 26,032 & 24,552 & 5.69\% & 1,748 & 93.29\% \\
\bottomrule
\end{tabular*}
\end{center}

\subsection{Category-wise Retention}
\label{app:construction-retention}
\label{app:category-retention}

Table~\ref{tab:A4-category-retention} reports category-wise construction retention, complementing the final L1/L2 counts shown in the main paper.

\begin{table}[!htbp]
\centering
\small
\caption{Category-wise retention across benchmark construction stages.}
\label{tab:A4-category-retention}
\renewcommand{\arraystretch}{1.10}
\begin{tabularx}{0.90\textwidth}{@{}>{\raggedright\arraybackslash}X>{\centering\arraybackslash}p{0.17\textwidth}>{\centering\arraybackslash}p{0.17\textwidth}>{\centering\arraybackslash}p{0.17\textwidth}@{}}
\toprule
\AppendixTableHead{Category} & \AppendixTableHead{L0$\rightarrow$L1} & \AppendixTableHead{L1$\rightarrow$L2} & \AppendixTableHead{L0$\rightarrow$L2} \\
\midrule
Time-lagged & 77.44\% & 76.87\% & 59.53\% \\
Synthetic / fictional & 66.24\% & 38.64\% & 25.59\% \\
Cultural blindspots & 61.68\% & 27.71\% & 17.09\% \\
Fine-grained / texture & 47.96\% & 50.26\% & 24.11\% \\
Environmental shift & 67.09\% & 49.66\% & 33.31\% \\
Viewpoint / scale & 52.65\% & 46.50\% & 24.48\% \\
Typographic conflict & 76.32\% & 33.01\% & 25.19\% \\
\bottomrule
\end{tabularx}
\end{table}

\subsection{Qualitative Benchmark Examples}
\label{app:construction-examples}
\label{app:qualitative-examples}

\begin{figure}[p]
\centering
\includegraphics[width=\linewidth,height=0.85\textheight,keepaspectratio]{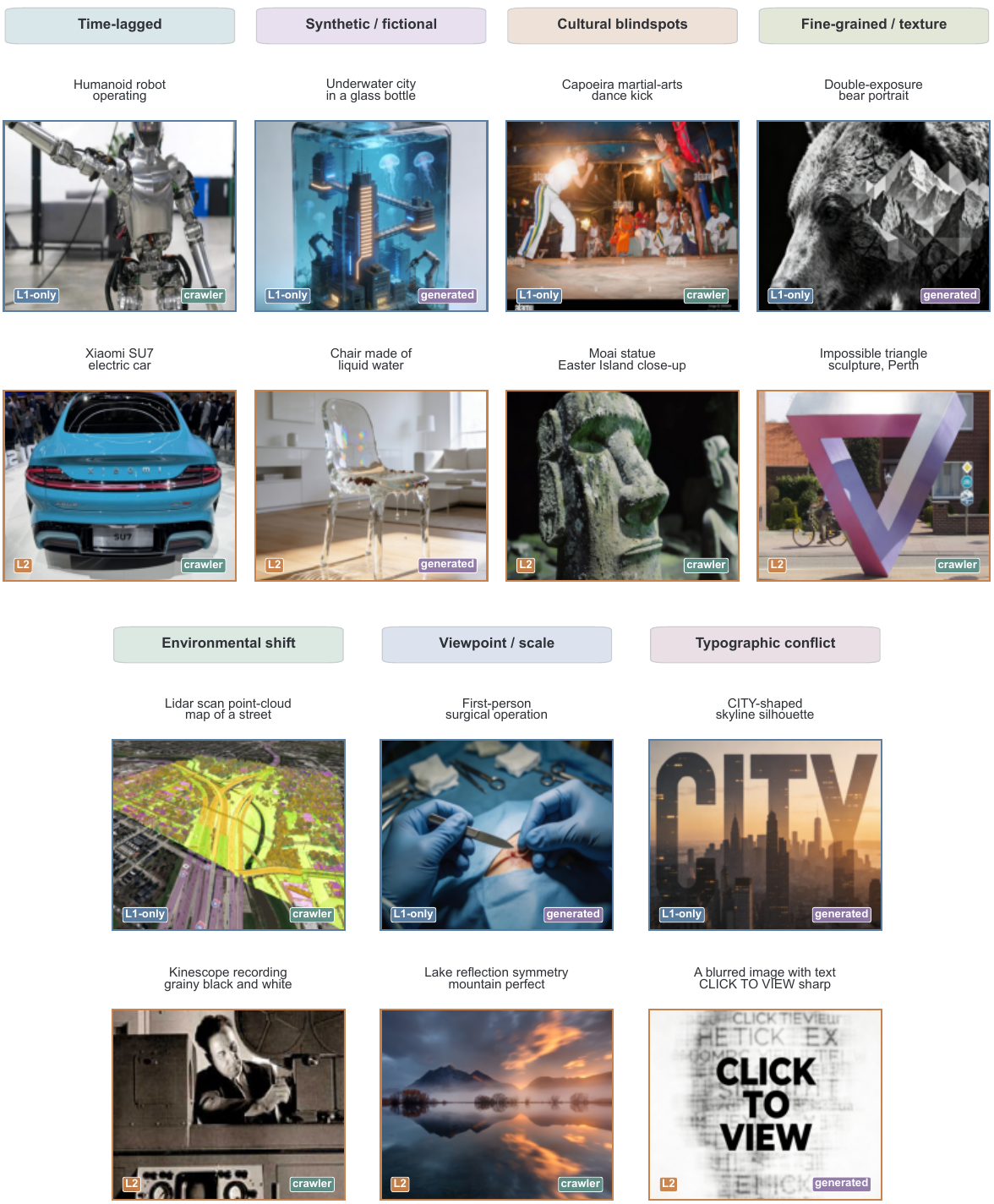}
\caption{Qualitative coverage across the seven benchmark categories. Within each category block, the upper sample is L1-only and the lower sample is L2; labels show the assigned target concept, benchmark layer, and image source.}
\label{fig:F1-real-examples}
\end{figure}

The manual technical review used during benchmark construction is described in Section~4.1. Figure~\ref{fig:F1-real-examples} shows representative L1-only and L2 examples across all seven benchmark categories.

\FloatBarrier

\section{Evaluation Theory}
\label{app:evaluation-theory}
\label{app:evaluation-protocol-details}
\setcounter{table}{0}
\setcounter{figure}{0}

\subsection{Finite-sample Background Exceedance Control}
\label{app:finite-sample}
\label{app:finite-sample-control}

For completeness, let
\[
Z_{m,i}=\max_{p\in P}s_m(X_i,p),
\]
and let $Z_{m,(k)}$ be the $k$th ordered calibration score, with $k=\lceil(n+1)(1-\alpha)\rceil$.

\paragraph{Theorem B1 (Finite-sample background exceedance control).}

Fix a model $m$ and target-prompt bank $P$. Let $X_1,\ldots,X_n,X_{n+1}$ be exchangeable background images and define
\[
Z_{m,i}=\max_{p\in P}s_m(X_i,p).
\]
Let $Z_{m,(1)}\le\cdots\le Z_{m,(n)}$ denote the ordered calibration scores. For $\alpha\in(0,1)$, set
\[
k=\left\lceil(n+1)(1-\alpha)\right\rceil,
\qquad
\widehat{\tau}_m=
\begin{cases}
Z_{m,(k)}, & k\le n,\\
+\infty, & k=n+1.
\end{cases}
\]
Then
\[
\Pr\!\left(\max_{p\in P}s_m(X_{n+1},p)>\widehat{\tau}_m\right)
\le \frac{n+1-k}{n+1}\le\alpha.
\]
When the scores have no ties almost surely, the first inequality is an equality. The result applies to the calibration distribution under exchangeability and does not require independence among scores for different target prompts.

\paragraph{Proof.}

Suppress the model index. Since the same fixed function
\[
T(x)=\max_{p\in P}s_m(x,p)
\]
is applied to exchangeable images, $Z_1,\ldots,Z_n,Z_{n+1}$ are exchangeable. Without ties, the rank of $Z_{n+1}$ among the $n+1$ scores is uniform on $\{1,\ldots,n+1\}$. The event $Z_{n+1}>\widehat{\tau}_m$ can occur only when this rank exceeds $k$, and hence
\[
\Pr\!\left(Z_{n+1}>\widehat{\tau}_m\right)
=\frac{n+1-k}{n+1}\le\alpha.
\]
With ties, strict exceedance can only reduce this probability. If $k=n+1$, then $\widehat{\tau}_m=+\infty$ and the exceedance probability is zero. $\square$

\subsection{Invariance to Strictly Increasing Score Transformations}
\label{app:score-invariance}
\label{app:monotone-invariance}

\paragraph{Corollary B1 (Invariance to strictly increasing score transformations).}

Let $g_m$ be any strictly increasing function and define
\[
\widetilde{s}_m(x,p)=g_m\!\left(s_m(x,p)\right).
\]
If calibration and evaluation are both performed using $\widetilde{s}_m$, then every image--concept recognition decision is unchanged. In particular,
\[
\widetilde Z_{m,i}
=\max_{p\in P}g_m\!\left(s_m(X_i,p)\right)
=g_m(Z_{m,i}),
\qquad
\widetilde{\tau}_m=g_m(\widehat{\tau}_m),
\]
and therefore
\[
\widetilde{s}_m(x,p)>\widetilde{\tau}_m
\quad\Longleftrightarrow\quad
s_m(x,p)>\widehat{\tau}_m.
\]

\paragraph{Proof.}

Strict monotonicity gives
\[
\max_{p\in P}g_m\!\left(s_m(X_i,p)\right)
=g_m\!\left(\max_{p\in P}s_m(X_i,p)\right)=g_m(Z_{m,i}).
\]
Thus the ordering of calibration scores is unchanged and the transformed $k$th order statistic is $g_m(\widehat{\tau}_m)$. The displayed equivalence follows, so every recognition decision and every reported recognition rate is unchanged. $\square$

\FloatBarrier

\section{Full Cross-model Results}
\label{app:full-results}
\label{app:full-cross-model-results}
\label{app:full-run-cross-model}
\label{app:category-slice-structure}
\setcounter{table}{0}
\setcounter{figure}{0}

\subsection{Model Registry and Overall Results}
\label{app:model-registry}

Table~\ref{tab:C1-model-registry-results} reports the fixed 37-model registry, model-specific $Q_{95,m}$, and overall L1/L2 recognition rates. All evaluated model-layer runs completed without invalid benchmark entries.

\begingroup
\scriptsize
\setlength{\tabcolsep}{4.2pt}
\renewcommand{\arraystretch}{1.07}
\setlength{\LTleft}{\fill}
\setlength{\LTright}{\fill}
\begin{longtable}{@{}>{\raggedright\arraybackslash}p{3.35cm}c>{\raggedright\arraybackslash}p{5.35cm}ccc@{}}
\caption{Model registry and overall target-concept recognition results.}\label{tab:C1-model-registry-results}\\
\toprule
\AppendixTableHead{Variant} & \AppendixTableHead{Year} & \AppendixTableHead{Backbone} & \AppendixTableHead{$Q_{95,m}$} & \AppendixTableHead{L1} & \AppendixTableHead{L2} \\
\midrule
\endfirsthead
\multicolumn{6}{l}{\tablename\ \thetable\ continued}\\
\toprule
\AppendixTableHead{Variant} & \AppendixTableHead{Year} & \AppendixTableHead{Backbone} & \AppendixTableHead{$Q_{95,m}$} & \AppendixTableHead{L1} & \AppendixTableHead{L2} \\
\midrule
\endhead
\midrule
\multicolumn{6}{r}{Continued on next page}\\
\endfoot
\bottomrule
\endlastfoot
\multicolumn{6}{@{}l@{}}{\AppendixFamilyTag{ConvNeXt-CLIP}}\\*
Base-W & 2022 & convnext\_base\_w & 0.3750 & 0.759 & 0.516 \\*
Base-W AugReg & 2022 & convnext\_base\_w & 0.3454 & 0.767 & 0.536 \\*
Large-D AugReg & 2022 & convnext\_large\_d & 0.2279 & 0.778 & 0.557 \\*
Large-D-320 FT & 2022 & convnext\_large\_d\_320 & 0.2273 & 0.788 & 0.575 \\
\multicolumn{6}{@{}l@{}}{\AppendixFamilyTag{DFN-CLIP}}\\*
B/16 & 2024 & ViT-B-16 & 0.2444 & 0.757 & 0.528 \\*
B/16 QuickGELU & 2024 & ViT-B-16-quickgelu & 0.2815 & 0.707 & 0.452 \\*
L/14 & 2024 & ViT-L-14 & 0.2186 & 0.772 & 0.561 \\
\multicolumn{6}{@{}l@{}}{\AppendixFamilyTag{DataComp-CLIP}}\\*
B/16 L & 2023 & ViT-B-16 & 0.3430 & 0.691 & 0.424 \\*
B/16 XL & 2023 & ViT-B-16 & 0.2546 & 0.794 & 0.583 \\*
B/32 M & 2023 & ViT-B-32 & 0.4175 & 0.251 & 0.106 \\*
B/32 S & 2023 & ViT-B-32 & 0.5797 & 0.013 & 0.005 \\
\multicolumn{6}{@{}l@{}}{\AppendixFamilyTag{EVA-CLIP}}\\*
EVA02-B/16 & 2023 & EVA02-B-16 & 0.2012 & 0.747 & 0.500 \\*
EVA02-L/14 & 2023 & EVA02-L-14 & 0.2393 & 0.798 & 0.596 \\
\multicolumn{6}{@{}l@{}}{\AppendixFamilyTag{MetaCLIP}}\\*
B/16 400M & 2023 & ViT-B-16 & 0.3252 & 0.626 & 0.362 \\*
B/16 FullCC & 2023 & ViT-B-16 & 0.3411 & 0.790 & 0.583 \\*
B/16 QuickGELU 400M & 2023 & ViT-B-16-quickgelu & 0.3278 & 0.689 & 0.446 \\
\multicolumn{6}{@{}l@{}}{\AppendixFamilyTag{OpenAI-CLIP}}\\*
B/32 local & 2021 & ViT-B/32 & 0.2949 & 0.605 & 0.150 \\*
B/16 & 2021 & ViT-B-16 & 0.2951 & 0.668 & 0.374 \\*
B/16 QuickGELU & 2021 & ViT-B-16-quickgelu & 0.2964 & 0.669 & 0.358 \\*
B/32 & 2021 & ViT-B-32 & 0.3004 & 0.530 & 0.106 \\*
B/32 QuickGELU & 2021 & ViT-B-32-quickgelu & 0.2948 & 0.607 & 0.153 \\
\multicolumn{6}{@{}l@{}}{\AppendixFamilyTag{DataComp/CommonPool}}\\*
B/16 CLIP-filtered & 2023 & ViT-B-16 & 0.3549 & 0.623 & 0.337 \\*
B/16 image-filtered & 2023 & ViT-B-16 & 0.3166 & 0.702 & 0.443 \\*
B/16 text-filtered & 2023 & ViT-B-16 & 0.3205 & 0.671 & 0.403 \\
\multicolumn{6}{@{}l@{}}{\AppendixFamilyTag{OpenCLIP-LAION}}\\*
H/14 local & 2022 & ViT-H-14 & 0.2809 & 0.762 & 0.530 \\*
B/16 & 2022 & ViT-B-16 & 0.2632 & 0.722 & 0.457 \\*
B/32 E16 & 2022 & ViT-B-32 & 0.3104 & 0.683 & 0.391 \\*
B/32 S34B & 2022 & ViT-B-32 & 0.2867 & 0.702 & 0.419 \\*
RoBERTa-B/32 & 2022 & roberta-ViT-B-32 & 0.2452 & 0.666 & 0.377 \\
\multicolumn{6}{@{}l@{}}{\AppendixFamilyTag{SigLIP}}\\*
B/16 256 & 2023 & ViT-B-16-SigLIP-256 & 0.1060 & 0.666 & 0.450 \\*
B/16 384 & 2023 & ViT-B-16-SigLIP-384 & 0.1007 & 0.648 & 0.433 \\*
B/16 i18n-256 & 2023 & ViT-B-16-SigLIP-i18n-256 & 0.1034 & 0.672 & 0.447 \\*
B/16 default & 2023 & ViT-B-16-SigLIP & 0.0998 & 0.677 & 0.464 \\
\multicolumn{6}{@{}l@{}}{\AppendixFamilyTag{SigLIP2}}\\*
B/16 256 & 2025 & ViT-B-16-SigLIP2-256 & 0.1296 & 0.737 & 0.535 \\*
B/16 384 & 2025 & ViT-B-16-SigLIP2-384 & 0.1318 & 0.724 & 0.527 \\*
B/16 512 & 2025 & ViT-B-16-SigLIP2-512 & 0.1365 & 0.695 & 0.489 \\*
B/16 default & 2025 & ViT-B-16-SigLIP2 & 0.1295 & 0.745 & 0.544 \\
\end{longtable}
\endgroup

\clearpage
\subsection{Full Category-level Results on L1}
\label{app:l1-category-results}

\begingroup
\small
\setlength{\tabcolsep}{2.5pt}
\renewcommand{\arraystretch}{0.98}
\begin{longtable}{@{}>{\raggedright\arraybackslash}p{3.3cm}rrrrrrrr@{}}
\caption{Per-model target-concept recognition rates by concept category on L1, grouped by model family. Category headers are abbreviated for layout; the full names are given below the table. Short variant labels are used for readability.}\label{tab:L5-l1-category-pass-rates}\\
\toprule
\AppendixTableHead{Variant} & \AppendixTableHead{Time} & \AppendixTableHead{Synth.} & \AppendixTableHead{Culture} & \AppendixTableHead{Fine/tex.} & \AppendixTableHead{Environ.} & \AppendixTableHead{View/scale} & \AppendixTableHead{Typogr.} & \AppendixTableHead{Overall} \\
\midrule
\endfirsthead
\multicolumn{9}{l}{\tablename\ \thetable\ continued}\\
\toprule
\AppendixTableHead{Variant} & \AppendixTableHead{Time} & \AppendixTableHead{Synth.} & \AppendixTableHead{Culture} & \AppendixTableHead{Fine/tex.} & \AppendixTableHead{Environ.} & \AppendixTableHead{View/scale} & \AppendixTableHead{Typogr.} & \AppendixTableHead{Overall} \\
\midrule
\endhead
\midrule
\multicolumn{9}{r}{Continued on next page}\\
\endfoot
\bottomrule
\endlastfoot
\multicolumn{9}{@{}l@{}}{\AppendixFamilyBandRow{ConvNeXt-CLIP}}\\*
Base-W & 0.499 & 0.922 & 0.831 & 0.632 & 0.758 & 0.786 & 0.865 & 0.759 \\*
Base-W AugReg & 0.539 & 0.941 & 0.806 & 0.631 & 0.765 & 0.779 & 0.871 & 0.767 \\*
Large-D AugReg & 0.541 & 0.961 & 0.826 & 0.638 & 0.772 & 0.796 & 0.878 & 0.778 \\*
Large-D-320 FT & 0.567 & 0.960 & 0.842 & 0.639 & 0.776 & 0.812 & 0.883 & 0.788 \\
\multicolumn{9}{@{}l@{}}{\AppendixFamilyBandRow{DFN-CLIP}}\\*
B/16 & 0.452 & 0.937 & 0.825 & 0.636 & 0.774 & 0.789 & 0.871 & 0.757 \\*
B/16 QuickGELU & 0.388 & 0.881 & 0.778 & 0.607 & 0.717 & 0.733 & 0.837 & 0.707 \\*
L/14 & 0.509 & 0.950 & 0.828 & 0.652 & 0.786 & 0.810 & 0.858 & 0.772 \\
\multicolumn{9}{@{}l@{}}{\AppendixFamilyBandRow{DataComp-CLIP}}\\*
B/16 L & 0.431 & 0.850 & 0.731 & 0.607 & 0.696 & 0.688 & 0.815 & 0.691 \\*
B/16 XL & 0.560 & 0.943 & 0.889 & 0.666 & 0.786 & 0.821 & 0.880 & 0.794 \\*
B/32 M & 0.070 & 0.279 & 0.204 & 0.258 & 0.277 & 0.270 & 0.384 & 0.251 \\*
B/32 S & 0.001 & 0.011 & 0.005 & 0.020 & 0.014 & 0.011 & 0.028 & 0.013 \\
\multicolumn{9}{@{}l@{}}{\AppendixFamilyBandRow{EVA-CLIP}}\\*
EVA02-B/16 & 0.451 & 0.923 & 0.835 & 0.625 & 0.743 & 0.792 & 0.849 & 0.747 \\*
EVA02-L/14 & 0.571 & 0.970 & 0.871 & 0.663 & 0.791 & 0.825 & 0.877 & 0.798 \\
\multicolumn{9}{@{}l@{}}{\AppendixFamilyBandRow{MetaCLIP}}\\*
B/16 400M & 0.372 & 0.857 & 0.546 & 0.560 & 0.639 & 0.582 & 0.777 & 0.626 \\*
B/16 FullCC & 0.573 & 0.964 & 0.810 & 0.666 & 0.777 & 0.812 & 0.891 & 0.790 \\*
B/16 QuickGELU 400M & 0.467 & 0.896 & 0.631 & 0.593 & 0.694 & 0.692 & 0.807 & 0.689 \\
\multicolumn{9}{@{}l@{}}{\AppendixFamilyBandRow{OpenAI-CLIP}}\\*
B/32 local & 0.300 & 0.694 & 0.780 & 0.544 & 0.569 & 0.621 & 0.743 & 0.605 \\*
B/16 & 0.329 & 0.857 & 0.807 & 0.611 & 0.647 & 0.637 & 0.801 & 0.668 \\*
B/16 QuickGELU & 0.373 & 0.802 & 0.843 & 0.624 & 0.636 & 0.656 & 0.775 & 0.669 \\*
B/32 & 0.218 & 0.663 & 0.680 & 0.494 & 0.512 & 0.481 & 0.674 & 0.530 \\*
B/32 QuickGELU & 0.302 & 0.697 & 0.781 & 0.545 & 0.570 & 0.622 & 0.745 & 0.607 \\
\multicolumn{9}{@{}l@{}}{\AppendixFamilyBandRow{DataComp/CommonPool}}\\*
B/16 CLIP-filtered & 0.395 & 0.731 & 0.690 & 0.573 & 0.592 & 0.560 & 0.790 & 0.623 \\*
B/16 image-filtered & 0.434 & 0.832 & 0.751 & 0.602 & 0.708 & 0.726 & 0.839 & 0.702 \\*
B/16 text-filtered & 0.374 & 0.828 & 0.684 & 0.584 & 0.702 & 0.687 & 0.814 & 0.671 \\
\multicolumn{9}{@{}l@{}}{\AppendixFamilyBandRow{OpenCLIP-LAION}}\\*
H/14 local & 0.506 & 0.942 & 0.837 & 0.631 & 0.762 & 0.769 & 0.866 & 0.762 \\*
B/16 & 0.430 & 0.906 & 0.770 & 0.614 & 0.738 & 0.738 & 0.839 & 0.722 \\*
B/32 E16 & 0.393 & 0.834 & 0.723 & 0.578 & 0.701 & 0.707 & 0.818 & 0.683 \\*
B/32 S34B & 0.417 & 0.864 & 0.756 & 0.594 & 0.713 & 0.716 & 0.835 & 0.702 \\*
RoBERTa-B/32 & 0.382 & 0.846 & 0.678 & 0.543 & 0.688 & 0.670 & 0.811 & 0.666 \\
\multicolumn{9}{@{}l@{}}{\AppendixFamilyBandRow{SigLIP}}\\*
B/16 256 & 0.465 & 0.906 & 0.534 & 0.558 & 0.695 & 0.653 & 0.782 & 0.666 \\*
B/16 384 & 0.447 & 0.878 & 0.518 & 0.553 & 0.683 & 0.617 & 0.772 & 0.648 \\*
B/16 i18n-256 & 0.470 & 0.901 & 0.593 & 0.566 & 0.680 & 0.636 & 0.797 & 0.672 \\*
B/16 default & 0.459 & 0.935 & 0.546 & 0.560 & 0.702 & 0.673 & 0.796 & 0.677 \\
\multicolumn{9}{@{}l@{}}{\AppendixFamilyBandRow{SigLIP2}}\\*
B/16 256 & 0.556 & 0.947 & 0.690 & 0.603 & 0.734 & 0.712 & 0.851 & 0.737 \\*
B/16 384 & 0.543 & 0.943 & 0.641 & 0.596 & 0.735 & 0.698 & 0.840 & 0.724 \\*
B/16 512 & 0.510 & 0.923 & 0.608 & 0.579 & 0.705 & 0.648 & 0.821 & 0.695 \\*
B/16 default & 0.560 & 0.952 & 0.709 & 0.601 & 0.742 & 0.735 & 0.856 & 0.745 \\
\end{longtable}
\noindent Time = Time-lagged; Synth. = Synthetic / fictional; Culture = Cultural blindspots; Fine/tex. = Fine-grained / texture; Environ. = Environmental shift; View/scale = Viewpoint / scale; Typogr. = Typographic conflict.
\endgroup

\clearpage
\subsection{Full Category-level Results on L2}
\label{app:l2-category-results}

\begingroup
\small
\setlength{\tabcolsep}{2.5pt}
\renewcommand{\arraystretch}{0.98}
\setlength{\LTpost}{0pt}
\begin{longtable}{@{}>{\raggedright\arraybackslash}p{3.3cm}rrrrrrrr@{}}
\caption{Per-model target-concept recognition rates by concept category on L2, grouped by model family. Category headers are abbreviated for layout; the full names are given below the table. Short variant labels are used for readability.}\label{tab:L6-l2-category-pass-rates}\\
\toprule
\AppendixTableHead{Variant} & \AppendixTableHead{Time} & \AppendixTableHead{Synth.} & \AppendixTableHead{Culture} & \AppendixTableHead{Fine/tex.} & \AppendixTableHead{Environ.} & \AppendixTableHead{View/scale} & \AppendixTableHead{Typogr.} & \AppendixTableHead{Overall} \\
\midrule
\endfirsthead
\multicolumn{9}{l}{\tablename\ \thetable\ continued}\\
\toprule
\AppendixTableHead{Variant} & \AppendixTableHead{Time} & \AppendixTableHead{Synth.} & \AppendixTableHead{Culture} & \AppendixTableHead{Fine/tex.} & \AppendixTableHead{Environ.} & \AppendixTableHead{View/scale} & \AppendixTableHead{Typogr.} & \AppendixTableHead{Overall} \\
\midrule
\endhead
\midrule
\multicolumn{9}{r}{Continued on next page}\\
\endfoot
\bottomrule
\endlastfoot
\multicolumn{9}{@{}l@{}}{\AppendixFamilyBandRow{ConvNeXt-CLIP}}\\*
Base-W & 0.373 & 0.807 & 0.516 & 0.295 & 0.526 & 0.582 & 0.645 & 0.516 \\*
Base-W AugReg & 0.416 & 0.854 & 0.485 & 0.299 & 0.544 & 0.582 & 0.653 & 0.536 \\*
Large-D AugReg & 0.417 & 0.903 & 0.526 & 0.312 & 0.555 & 0.620 & 0.677 & 0.557 \\*
Large-D-320 FT & 0.448 & 0.906 & 0.568 & 0.314 & 0.561 & 0.653 & 0.687 & 0.575 \\
\multicolumn{9}{@{}l@{}}{\AppendixFamilyBandRow{DFN-CLIP}}\\*
B/16 & 0.340 & 0.848 & 0.550 & 0.317 & 0.563 & 0.607 & 0.663 & 0.528 \\*
B/16 QuickGELU & 0.279 & 0.736 & 0.479 & 0.264 & 0.483 & 0.508 & 0.594 & 0.452 \\*
L/14 & 0.398 & 0.893 & 0.567 & 0.341 & 0.585 & 0.662 & 0.637 & 0.561 \\
\multicolumn{9}{@{}l@{}}{\AppendixFamilyBandRow{DataComp-CLIP}}\\*
B/16 L & 0.305 & 0.661 & 0.413 & 0.259 & 0.431 & 0.441 & 0.564 & 0.424 \\*
B/16 XL & 0.448 & 0.868 & 0.655 & 0.356 & 0.579 & 0.658 & 0.675 & 0.583 \\*
B/32 M & 0.045 & 0.156 & 0.058 & 0.056 & 0.120 & 0.136 & 0.201 & 0.106 \\*
B/32 S & 0.001 & 0.007 & 0.003 & 0.005 & 0.001 & 0.005 & 0.015 & 0.005 \\
\multicolumn{9}{@{}l@{}}{\AppendixFamilyBandRow{EVA-CLIP}}\\*
EVA02-B/16 & 0.321 & 0.812 & 0.535 & 0.289 & 0.513 & 0.602 & 0.620 & 0.500 \\*
EVA02-L/14 & 0.460 & 0.929 & 0.650 & 0.348 & 0.589 & 0.665 & 0.678 & 0.596 \\
\multicolumn{9}{@{}l@{}}{\AppendixFamilyBandRow{MetaCLIP}}\\*
B/16 400M & 0.256 & 0.680 & 0.229 & 0.211 & 0.348 & 0.375 & 0.483 & 0.362 \\*
B/16 FullCC & 0.467 & 0.909 & 0.523 & 0.360 & 0.566 & 0.637 & 0.698 & 0.583 \\*
B/16 QuickGELU 400M & 0.350 & 0.756 & 0.312 & 0.260 & 0.436 & 0.498 & 0.537 & 0.446 \\
\multicolumn{9}{@{}l@{}}{\AppendixFamilyBandRow{OpenAI-CLIP}}\\*
B/32 local & 0.090 & 0.207 & 0.206 & 0.092 & 0.131 & 0.185 & 0.223 & 0.150 \\*
B/16 & 0.184 & 0.674 & 0.464 & 0.269 & 0.358 & 0.386 & 0.532 & 0.374 \\*
B/16 QuickGELU & 0.220 & 0.557 & 0.508 & 0.275 & 0.328 & 0.380 & 0.462 & 0.358 \\*
B/32 & 0.049 & 0.219 & 0.130 & 0.068 & 0.103 & 0.070 & 0.179 & 0.106 \\*
B/32 QuickGELU & 0.092 & 0.215 & 0.211 & 0.094 & 0.135 & 0.188 & 0.227 & 0.153 \\
\multicolumn{9}{@{}l@{}}{\AppendixFamilyBandRow{DataComp/CommonPool}}\\*
B/16 CLIP-filtered & 0.266 & 0.451 & 0.336 & 0.219 & 0.292 & 0.331 & 0.527 & 0.337 \\*
B/16 image-filtered & 0.322 & 0.626 & 0.452 & 0.262 & 0.445 & 0.505 & 0.601 & 0.443 \\*
B/16 text-filtered & 0.260 & 0.639 & 0.382 & 0.247 & 0.444 & 0.429 & 0.547 & 0.403 \\
\multicolumn{9}{@{}l@{}}{\AppendixFamilyBandRow{OpenCLIP-LAION}}\\*
H/14 local & 0.381 & 0.860 & 0.545 & 0.302 & 0.542 & 0.568 & 0.652 & 0.530 \\*
B/16 & 0.293 & 0.770 & 0.409 & 0.276 & 0.489 & 0.507 & 0.591 & 0.457 \\*
B/32 E16 & 0.255 & 0.615 & 0.313 & 0.209 & 0.431 & 0.461 & 0.536 & 0.391 \\*
B/32 S34B & 0.279 & 0.671 & 0.361 & 0.239 & 0.445 & 0.466 & 0.572 & 0.419 \\*
RoBERTa-B/32 & 0.250 & 0.632 & 0.286 & 0.175 & 0.412 & 0.427 & 0.527 & 0.377 \\
\multicolumn{9}{@{}l@{}}{\AppendixFamilyBandRow{SigLIP}}\\*
B/16 256 & 0.356 & 0.796 & 0.279 & 0.236 & 0.451 & 0.528 & 0.510 & 0.450 \\*
B/16 384 & 0.339 & 0.749 & 0.293 & 0.241 & 0.444 & 0.484 & 0.502 & 0.433 \\*
B/16 i18n-256 & 0.361 & 0.785 & 0.308 & 0.247 & 0.435 & 0.482 & 0.531 & 0.447 \\*
B/16 default & 0.348 & 0.855 & 0.300 & 0.246 & 0.458 & 0.542 & 0.537 & 0.464 \\
\multicolumn{9}{@{}l@{}}{\AppendixFamilyBandRow{SigLIP2}}\\*
B/16 256 & 0.452 & 0.872 & 0.411 & 0.285 & 0.508 & 0.588 & 0.641 & 0.535 \\*
B/16 384 & 0.441 & 0.865 & 0.370 & 0.289 & 0.520 & 0.577 & 0.629 & 0.527 \\*
B/16 512 & 0.405 & 0.821 & 0.339 & 0.272 & 0.482 & 0.517 & 0.590 & 0.489 \\*
B/16 default & 0.456 & 0.884 & 0.419 & 0.280 & 0.523 & 0.605 & 0.654 & 0.544 \\
\end{longtable}
\noindent Time = Time-lagged; Synth. = Synthetic / fictional; Culture = Cultural blindspots; Fine/tex. = Fine-grained / texture; Environ. = Environmental shift; View/scale = Viewpoint / scale; Typogr. = Typographic conflict.
\endgroup

\subsection{Cross-model Category Summary}
\label{app:category-summary}

Across the 259 paired model--category slices, 258 have lower recognition on L2 than on L1. The sole exception is a near tie for DataComp-CLIP B/32 S on time-lagged concepts.

\begingroup
\makeatletter\def\@captype{table}\makeatother
\centering
\footnotesize\setlength{\tabcolsep}{2pt}
\caption{Cross-model category concept-recognition summary.}
\label{tab:K1-category-cross-model}
\renewcommand{\arraystretch}{1.08}
\begin{tabularx}{0.78\textwidth}{@{}>{\raggedright\arraybackslash}Xrrrr@{}}
\toprule
\AppendixTableHead{Category} & \AppendixTableHead{\shortstack{L1\\recognition}} & \AppendixTableHead{\shortstack{L2\\recognition}} & \AppendixTableHead{\shortstack{L1--L2\\difference}} & \AppendixTableHead{\shortstack{Models with\\L2$<$L1}} \\
\midrule
Time-lagged & 42.9\% & 30.9\% & 12.0 pp & 36/37 \\
Synthetic / fictional & 84.1\% & 68.9\% & 15.2 pp & 37/37 \\
Cultural blindspots & 69.9\% & 38.9\% & 31.0 pp & 37/37 \\
Fine-grained / texture & 57.4\% & 24.6\% & 32.8 pp & 37/37 \\
Environmental shift & 67.5\% & 42.6\% & 24.9 pp & 37/37 \\
Viewpoint / scale & 67.5\% & 47.2\% & 20.3 pp & 37/37 \\
Typographic conflict & 79.0\% & 53.5\% & 25.5 pp & 37/37 \\
\bottomrule
\end{tabularx}
\par\endgroup

\subsection{Release-year and Variant Analyses}
\label{app:release-variant}

These descriptive tables compare release-year groups and selected variants; differences may reflect multiple model and training factors.

\par\noindent\begin{minipage}{\linewidth}\makeatletter\def\@captype{table}\makeatother
\centering
\caption{Model-weighted concept-recognition trends by release-year group. Rates and L1--L2 differences are percentages; year groups differ in family, data, objective, resolution, and backbone composition (Section~6.3).}
\label{tab:L7-year-trend}
\scriptsize
\setlength{\tabcolsep}{4pt}
\begin{tabular*}{0.84\textwidth}{@{\extracolsep{\fill}}lrrrrrrr@{}}
\toprule
\AppendixTableHead{Year} & \AppendixTableHead{Models} & \AppendixTableHead{\shortstack{Mean\\L1}} & \AppendixTableHead{\shortstack{Median\\L1}} & \AppendixTableHead{\shortstack{Mean\\L2}} & \AppendixTableHead{\shortstack{Median\\L2}} & \AppendixTableHead{\shortstack{Mean\\L1--L2 difference}} & \AppendixTableHead{\shortstack{Median\\L1--L2 difference}} \\
\midrule
2021 & 5 & 61.6 & 60.7 & 22.8 & 15.3 & 38.8 & 42.4 \\
2022 & 9 & 73.6 & 75.9 & 48.4 & 51.6 & 25.2 & 24.3 \\
2023 & 16 & 62.9 & 67.5 & 41.1 & 44.4 & 21.8 & 22.1 \\
2024 & 3 & 74.6 & 75.7 & 51.4 & 52.8 & 23.2 & 22.9 \\
2025 & 4 & 72.5 & 73.0 & 52.4 & 53.1 & 20.2 & 20.2 \\
\bottomrule
\end{tabular*}
\end{minipage}\par

\par\noindent\begin{minipage}{\linewidth}\makeatletter\def\@captype{table}\makeatother
\centering
\caption{Selected within-family variant comparisons. L2 target-concept recognition rates and differences are percentages. These comparisons are descriptive because training data, objectives, architectures, input resolutions, and implementations may also differ (Section~6.3).}
\label{tab:L8-backbone-comparisons}
\scriptsize
\setlength{\tabcolsep}{3pt}
\begin{tabularx}{\textwidth}{@{}>{\raggedright\arraybackslash}p{2.25cm}>{\raggedright\arraybackslash}p{2.10cm}>{\raggedright\arraybackslash}p{2.00cm}rrr>{\raggedright\arraybackslash}X@{}}
\toprule
\AppendixTableHead{Family} & \AppendixTableHead{\shortstack{Reference\\variant(s)}} & \AppendixTableHead{\shortstack{Compared\\variant(s)}} & \AppendixTableHead{Reference} & \AppendixTableHead{Compared} & \AppendixTableHead{Difference} & \AppendixTableHead{Caution} \\
\midrule
ConvNeXt-CLIP & Base-W variants & Large-D variants & 52.6 & 56.6 & 4.0 & Recipe and resolution differ. \\
DFN-CLIP & B/16 variants & L/14 & 49.0 & 56.1 & 7.1 & Training scale and implementation differ. \\
EVA-CLIP & B/16 & L/14 & 50.0 & 59.6 & 9.6 & Single-model comparison. \\
OpenAI CLIP & B/32 variants & B/16 variants & 13.6 & 36.6 & 23.0 & Patch size, resolution, and variants differ. \\
OpenCLIP-LAION & B/32 variants & B/16 & 39.5 & 45.7 & 6.1 & Training tags and data scale differ. \\
OpenCLIP-LAION & B/16 & H/14 & 45.7 & 53.0 & 7.3 & Training tags and data scale differ. \\
DataComp-CLIP & B/32 variants & B/16 variants & 5.5 & 50.4 & 44.8 & Highly confounded by data and training scale. \\
\bottomrule
\end{tabularx}
\end{minipage}\par

\par\noindent\begin{minipage}{\linewidth}\makeatletter\def\@captype{table}\makeatother
\centering
\caption{Model-weighted target-concept recognition by category for the 2021 and 2025 release-year groups at both benchmark levels. Rates and differences are percentages. Group means and differences are independently rounded to one decimal, with differences computed from the unrounded group means as 2025 minus 2021. The year groups are not matched causal samples (Section~6.4).}
\label{tab:L9-category-progress}
\footnotesize\setlength{\tabcolsep}{2pt}
\begin{minipage}[t]{0.49\linewidth}\vspace{0pt}\begin{tabularx}{\linewidth}{@{}Xrrr@{}}
\toprule
\multicolumn{4}{c}{(a) L1} \\
\AppendixTableHead{Category} & \AppendixTableHead{\shortstack{2021\\($n=5$)}} & \AppendixTableHead{\shortstack{2025\\($n=4$)}} & \AppendixTableHead{\shortstack{Difference\\(2025 $-$\\2021; pp)}} \\
\midrule
Time-lagged & 30.4 & 54.2 & $+23.8$ \\
Synthetic / fictional & 74.3 & 94.1 & $+19.9$ \\
Cultural blindspots & 77.8 & 66.2 & $-11.6$ \\
Fine-grained / texture & 56.4 & 59.5 & $+3.1$ \\
Environmental shift & 58.7 & 72.9 & $+14.2$ \\
Viewpoint / scale & 60.3 & 69.8 & $+9.5$ \\
Typographic conflict & 74.8 & 84.2 & $+9.5$ \\
\bottomrule\end{tabularx}\end{minipage}\hfill
\begin{minipage}[t]{0.49\linewidth}\vspace{0pt}\begin{tabularx}{\linewidth}{@{}Xrrr@{}}
\toprule
\multicolumn{4}{c}{(b) L2} \\
\AppendixTableHead{Category} & \AppendixTableHead{\shortstack{2021\\($n=5$)}} & \AppendixTableHead{\shortstack{2025\\($n=4$)}} & \AppendixTableHead{\shortstack{Difference\\(2025 $-$\\2021; pp)}} \\
\midrule
Time-lagged & 12.7 & 43.8 & $+31.1$ \\
Synthetic / fictional & 37.4 & 86.0 & $+48.6$ \\
Cultural blindspots & 30.4 & 38.5 & $+8.1$ \\
Fine-grained / texture & 16.0 & 28.2 & $+12.2$ \\
Environmental shift & 21.1 & 50.8 & $+29.7$ \\
Viewpoint / scale & 24.2 & 57.2 & $+33.0$ \\
Typographic conflict & 32.5 & 62.8 & $+30.4$ \\
\bottomrule
\end{tabularx}\end{minipage}
\end{minipage}\par

\par\noindent\begin{minipage}{\linewidth}\makeatletter\def\@captype{table}\makeatother
\centering
\caption{SigLIP-to-SigLIP2 concept-recognition comparison. Values are family means in percentages; category rows use L2 (Section~6.3).}
\label{tab:L10-siglip-comparison}
\small
\begin{tabular*}{0.62\textwidth}{@{\extracolsep{\fill}}lrrr@{}}
\toprule
\AppendixTableHead{Scope} & \AppendixTableHead{SigLIP} & \AppendixTableHead{SigLIP2} & \AppendixTableHead{Difference} \\
\midrule
L1 & 66.6 & 72.5 & 6.0 \\
L2 & 44.9 & 52.4 & 7.5 \\
Time-lagged & 35.1 & 43.8 & 8.7 \\
Synthetic / fictional & 79.6 & 86.0 & 6.4 \\
Cultural blindspots & 29.5 & 38.5 & 9.0 \\
Fine-grained / texture & 24.3 & 28.2 & 3.9 \\
Environmental shift & 44.7 & 50.8 & 6.1 \\
Viewpoint / scale & 50.9 & 57.2 & 6.3 \\
Typographic conflict & 52.0 & 62.8 & 10.8 \\
\bottomrule
\end{tabular*}
\end{minipage}\par

\clearpage
\subsection{Model--Category Recognition Maps}
\label{app:recognition-maps}

\begin{figure}[!htbp]
\centering
\includegraphics[width=\linewidth,height=0.85\textheight,keepaspectratio]{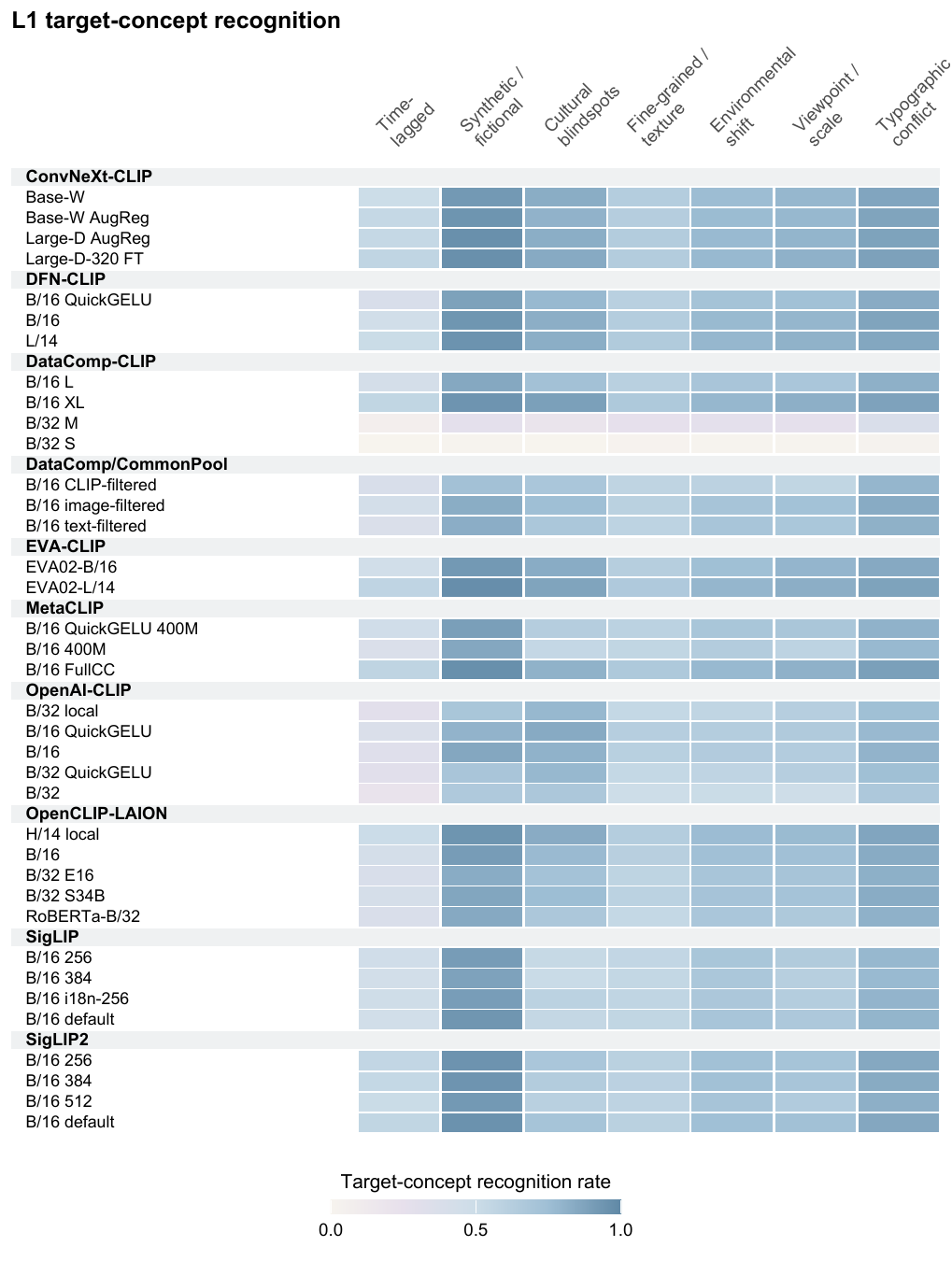}
\caption{L1 target-concept recognition by model and concept category. Models are grouped by family and use family-specific short variant labels.}
\label{fig:L1-l1-model-category-heatmap}
\end{figure}

\begin{figure}[!htbp]
\centering
\includegraphics[width=\linewidth,height=0.85\textheight,keepaspectratio]{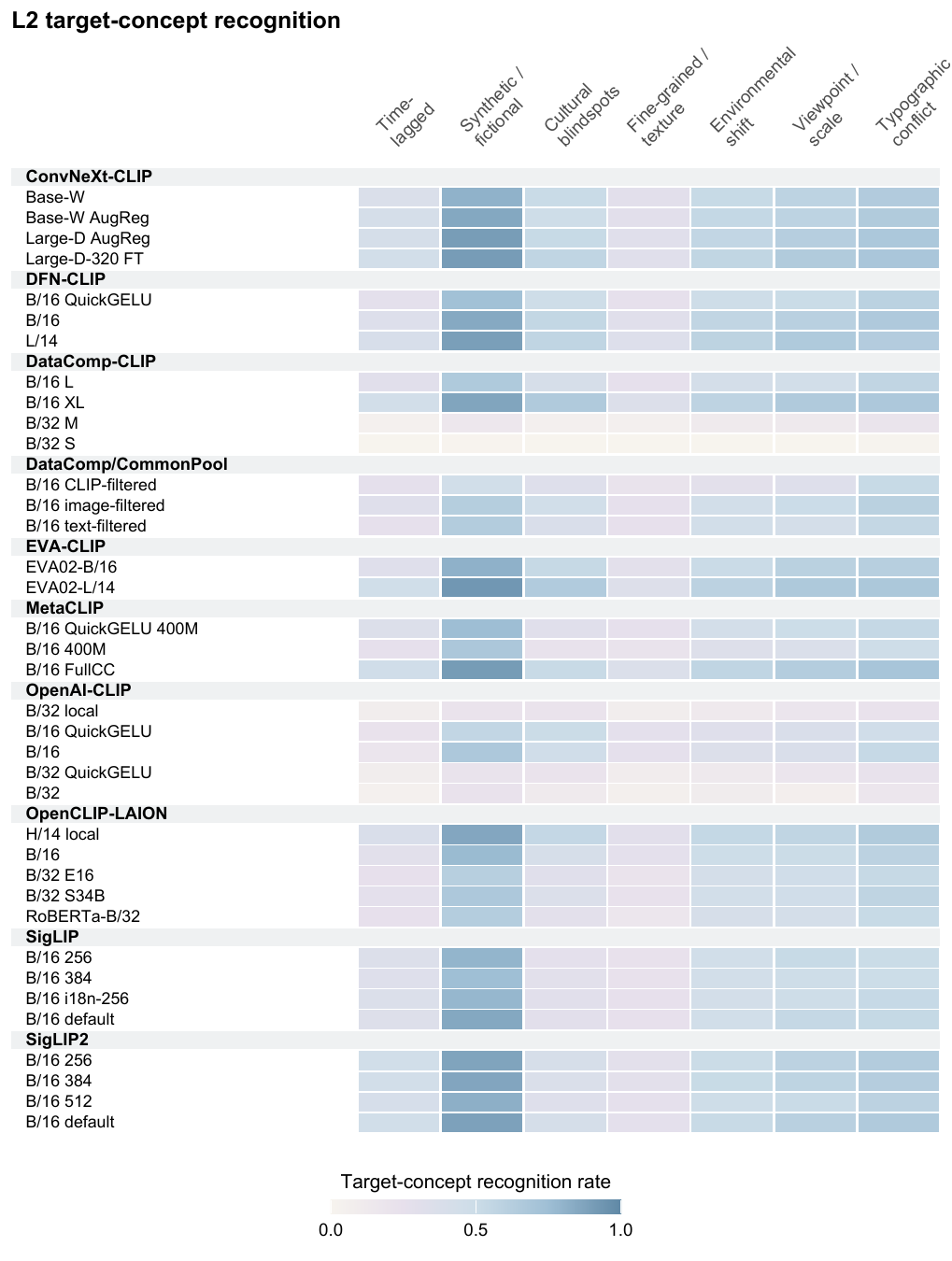}
\caption{L2 target-concept recognition by model and concept category. Models are grouped by family and use family-specific short variant labels.}
\label{fig:L2-l2-model-category-heatmap}
\end{figure}

Figures~\ref{fig:L1-l1-model-category-heatmap} and~\ref{fig:L2-l2-model-category-heatmap} provide the full model-by-category recognition patterns for L1 and L2.

\clearpage

\clearpage

\subsection{Concept-level Release-year Comparison}
\label{app:concept-year}

The 91 displayed concepts are selected from the L2-active concepts using 13 concepts per category. Selection is category-balanced and stratified by recognition level and release-year change while preferring concepts with sufficient image support. Ties are resolved deterministically, and the same 91 concepts are used for both release-year comparisons.

\begin{figure}[!htbp]
\centering
\includegraphics[width=\linewidth]{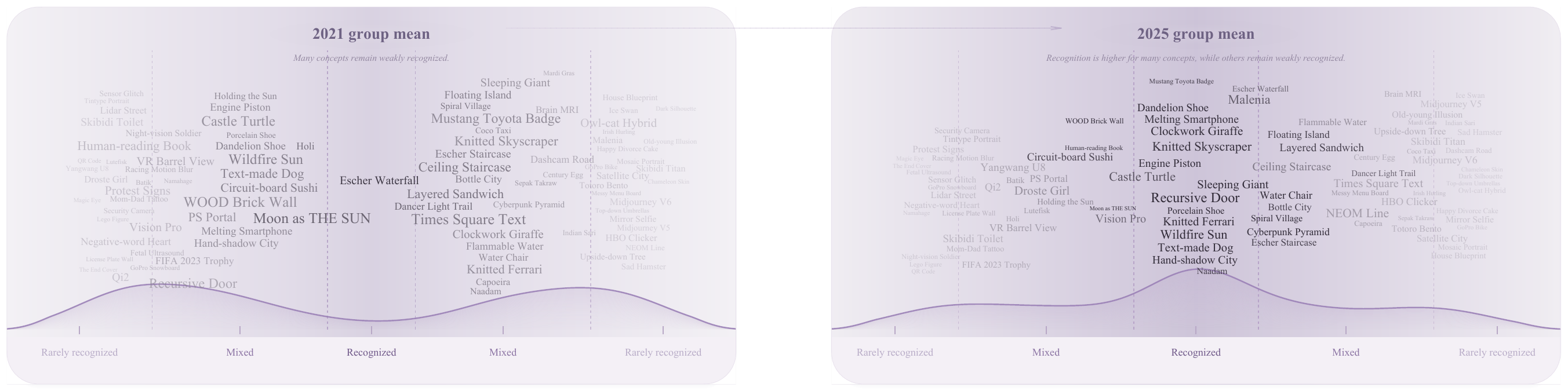}
\caption{Concept-level L2 recognition landscapes for the 2021 and 2025 release-year groups using the same fixed 91-concept subset.}
\label{fig:L4-l2-landscapes-2021-2025}
\end{figure}
\FloatBarrier

\section{Supporting Analyses}
\label{app:supporting}
\label{app:supporting-analyses}
\setcounter{table}{0}
\setcounter{figure}{0}

\subsection{Fixed-ID Reference Results}
\label{app:fixed-id}
\label{app:id-confidence-score-overlap}

Table~\ref{tab:I2-fixed-id-compact} applies the same fixed-reference metrics to Oxford-IIIT Pet, CIFAR-100, and STL-10. The results show that the measured separation depends on the selected ID reference.

\begin{table}[!htbp]
\centering
\scriptsize
\setlength{\tabcolsep}{2.6pt}
\renewcommand{\arraystretch}{1.06}
\caption{Fixed-ID results across reference datasets.}
\label{tab:I2-fixed-id-compact}
\begin{tabular*}{0.98\textwidth}{@{\extracolsep{\fill}}llrrrrrrrr@{}}
\toprule
\AppendixTableHead{ID baseline} & \AppendixTableHead{Layer} & \AppendixTableHead{$N_{ID}$} & \AppendixTableHead{$N_{benchmark}$} & \AppendixTableHead{Mean confidence} & \AppendixTableHead{AUROC} & \AppendixTableHead{FPR95} & \AppendixTableHead{AUPR-in} & \AppendixTableHead{AUPR-out} & \AppendixTableHead{ECE} \\
\midrule
Oxford-IIIT Pet & L1 & 3,669 & 30,752 & 0.263 & 0.830 & 0.650 & 0.420 & 0.974 & 0.187 \\
Oxford-IIIT Pet & L2 & 3,669 & 14,285 & 0.261 & 0.831 & 0.658 & 0.565 & 0.949 & 0.115 \\
CIFAR-100 & L1 & 10,000 & 30,752 & 0.263 & 0.395 & 0.949 & 0.217 & 0.667 & 0.183 \\
CIFAR-100 & L2 & 10,000 & 14,285 & 0.261 & 0.402 & 0.951 & 0.372 & 0.497 & 0.242 \\
STL-10 & L1 & 8,000 & 30,752 & 0.263 & 0.738 & 0.793 & 0.426 & 0.912 & 0.093 \\
STL-10 & L2 & 8,000 & 14,285 & 0.261 & 0.741 & 0.799 & 0.595 & 0.838 & 0.064 \\
\bottomrule
\end{tabular*}
\end{table}

\subsection{Limited-supervision Classification}
\label{app:limited-supervision}
\label{app:lightweight-supervision}

The task separates benchmark samples from a pooled common-category control within each layer and category. Training and test indices are disjoint, and results are averaged over five draws.

\begin{table}[!htbp]
\centering
\footnotesize
\renewcommand{\arraystretch}{1.06}
\caption{Limited-supervision benchmark-sample-versus-pooled-control classification protocol.}
\label{tab:D2-limited-supervision-protocol}
\begin{tabularx}{0.94\textwidth}{@{}>{\raggedright\arraybackslash}p{0.23\textwidth}>{\raggedright\arraybackslash}X@{}}
\toprule
\AppendixTableHead{Protocol element} & \AppendixTableHead{Specification} \\
\midrule
Encoder & CLIP ViT-B/32 with standard preprocessing and unit-normalized image features. \\
Task & Binary benchmark-sample-versus-pooled-control classification, run separately by L1/L2 layer and category. \\
Classifier & Logistic regression with the \texttt{liblinear} solver and 200 maximum iterations. \\
Positive class & Benchmark samples from the target layer/category slice. \\
Pooled control & Seven common-category control groups; no calibration/reference dataset is included. \\
Shots per class & 1, 3, 5, 10, 15, 20, 25, and 30. \\
Training sample & $n$ benchmark plus $n$ pooled-control examples, sampled without replacement. \\
Train/test separation & Test indices come from the class-specific remainder after training indices are removed; each test class is capped at 100 examples. \\
Repeated draws & Five per layer/category/shot; the reported value is their mean accuracy. \\
\bottomrule
\end{tabularx}
\end{table}

\FloatBarrier

\begin{table}[!htbp]
\centering
\small
\setlength{\tabcolsep}{3pt}
\renewcommand{\arraystretch}{1.03}
\caption{Layer-level limited-supervision classification means and category ranges.}
\label{tab:J2-fewshot-recovery}
\begin{tabular*}{0.58\textwidth}{@{\extracolsep{\fill}}lrrrr@{}}
\toprule
\AppendixTableHead{Layer} & \AppendixTableHead{Shots} & \shortstack{\AppendixTableHead{Mean}\\\AppendixTableHead{acc.}} & \shortstack{\AppendixTableHead{Cat.}\\\AppendixTableHead{min}} & \shortstack{\AppendixTableHead{Cat.}\\\AppendixTableHead{max}} \\
\midrule
L1 & 1 & 0.627 & 0.585 & 0.704 \\
L1 & 3 & 0.736 & 0.667 & 0.811 \\
L1 & 5 & 0.779 & 0.716 & 0.879 \\
L1 & 10 & 0.839 & 0.753 & 0.936 \\
L1 & 15 & 0.862 & 0.794 & 0.949 \\
L1 & 20 & 0.871 & 0.815 & 0.957 \\
L1 & 25 & 0.877 & 0.799 & 0.960 \\
L1 & 30 & 0.887 & 0.832 & 0.964 \\
L2 & 1 & 0.603 & 0.540 & 0.675 \\
L2 & 3 & 0.744 & 0.681 & 0.852 \\
L2 & 5 & 0.762 & 0.690 & 0.893 \\
L2 & 10 & 0.823 & 0.755 & 0.918 \\
L2 & 15 & 0.839 & 0.802 & 0.940 \\
L2 & 20 & 0.845 & 0.791 & 0.942 \\
L2 & 25 & 0.865 & 0.825 & 0.952 \\
L2 & 30 & 0.867 & 0.827 & 0.944 \\
\bottomrule
\end{tabular*}
\end{table}

\begin{figure}[!htbp]
\centering
\includegraphics[width=\textwidth]{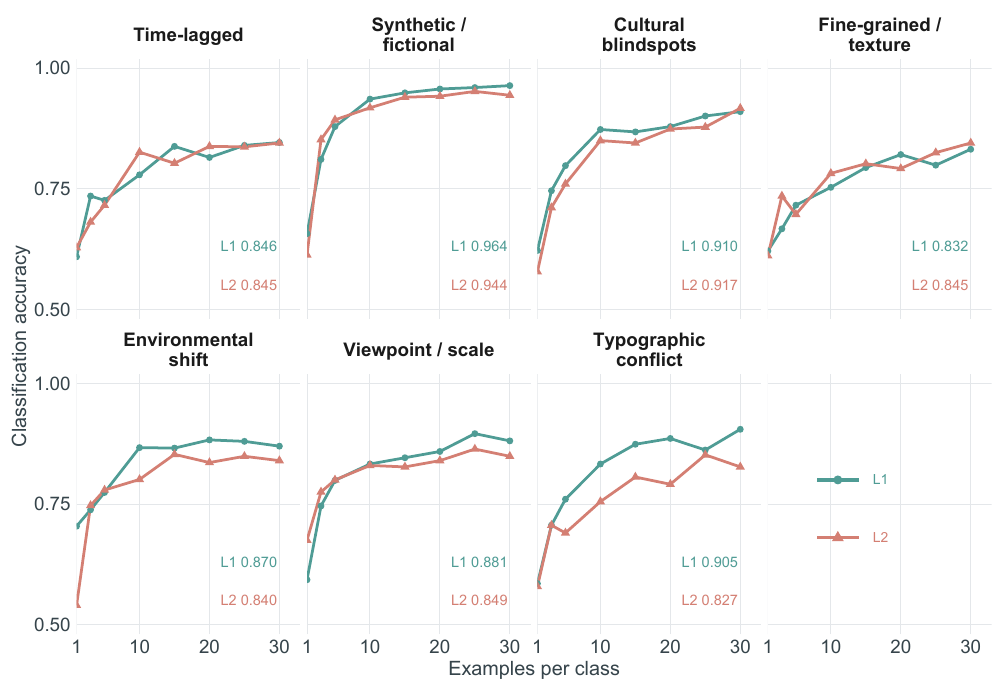}
\caption{Category-wise benchmark-sample-versus-pooled-control classification accuracy within each L1/L2 layer across all eight shot settings.}
\label{fig:J2-category-fewshot}
\end{figure}
\FloatBarrier

\subsection{Source-grouped Diagnostics}
\label{app:source-analysis}
\label{app:source-aware-diagnostics}

Here, crawler denotes web-search images. Within-category crawler/generated comparisons are available for Fine-grained / texture, Environmental shift, Viewpoint / scale, and Typographic conflict; the remaining categories are single-source.

\begin{figure}[!htbp]
\centering
\includegraphics[width=0.76\textwidth]{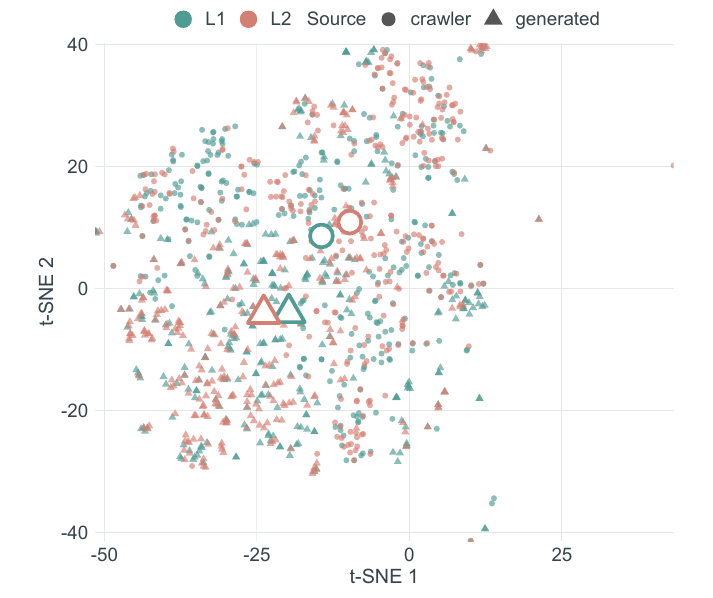}
\caption{Qualitative t-SNE view of the four L1/L2 layer--source groups using 300 samples per layer--source group. Color encodes layer, shape encodes source, and open markers show group centroids.}
\label{fig:M1-source-aware-tsne}
\end{figure}

Let $X_c$ and $X_g$ be the crawler and generated CLIP image-feature matrices with centroids $\mu_c$ and $\mu_g$. The centroid cosine distance is $1-\mu_c^\top\mu_g/(\lVert\mu_c\rVert_2\lVert\mu_g\rVert_2)$. The CLIP-feature Fr\'echet distance uses the corresponding feature means and covariances with $10^{-6}I$ stabilization. Target-concept cosine and confidence gaps use generated minus crawler means.

\begin{table}[!htbp]
\centering
\footnotesize
\setlength{\tabcolsep}{3.2pt}
\renewcommand{\arraystretch}{1.06}
\caption{Metric summary by image source for the fixed L1/L2 layers. ``Fr\'echet'' is the CLIP-feature Fr\'echet distance; the cosine gap is the target-concept cosine gap.}
\label{tab:M3-source-aware-metrics}
\begin{tabular*}{0.94\textwidth}{@{\extracolsep{\fill}}lrrrrrr@{}}
\toprule
\AppendixTableHead{Layer} & \AppendixTableHead{Crawler} & \AppendixTableHead{Generated} & \AppendixTableHead{Centroid cos. dist.} & \AppendixTableHead{Fr\'echet} & \AppendixTableHead{Target-concept cos. gap} & \AppendixTableHead{Conf. gap} \\
\midrule
L1 & 14,617 & 16,135 & 0.046 & 0.198 & 0.036 & 0.011 \\
L2 & 8,903 & 5,382 & 0.047 & 0.206 & 0.025 & 0.005 \\
\bottomrule
\end{tabular*}
\end{table}

\begingroup
\scriptsize
\setlength{\tabcolsep}{3pt}
\renewcommand{\arraystretch}{1.06}
\setlength\LTleft{\fill}
\setlength\LTright{\fill}
\begin{longtable}{@{}>{\raggedright\arraybackslash}p{0.7cm}>{\raggedright\arraybackslash}p{3.4cm}>{\raggedleft\arraybackslash}p{1.65cm}>{\raggedleft\arraybackslash}p{1.45cm}>{\raggedleft\arraybackslash}p{1.75cm}>{\raggedleft\arraybackslash}p{1.0cm}@{}}
\caption{Category-level source analyses for slices with both source types. Signed gaps use generated minus crawler means.}\label{tab:D5-category-source-metrics}\\
\toprule
\AppendixTableHead{Layer} & \AppendixTableHead{Category} & \shortstack{\AppendixTableHead{CLIP-feature}\\\AppendixTableHead{Fr\'echet dist.}} & \shortstack{\AppendixTableHead{Centroid}\\\AppendixTableHead{cos. dist.}} & \shortstack{\AppendixTableHead{Target-concept}\\\AppendixTableHead{cos. gap}} & \shortstack{\AppendixTableHead{Conf.}\\\AppendixTableHead{gap}} \\
\midrule
\endfirsthead
\multicolumn{6}{l}{\tablename\ \thetable\ continued}\\
\toprule
\AppendixTableHead{Layer} & \AppendixTableHead{Category} & \shortstack{\AppendixTableHead{CLIP-feature}\\\AppendixTableHead{Fr\'echet dist.}} & \shortstack{\AppendixTableHead{Centroid}\\\AppendixTableHead{cos. dist.}} & \shortstack{\AppendixTableHead{Target-concept}\\\AppendixTableHead{cos. gap}} & \shortstack{\AppendixTableHead{Conf.}\\\AppendixTableHead{gap}} \\
\midrule
\endhead
\midrule
\multicolumn{6}{r}{Continued on next page}\\
\endfoot
\bottomrule
\endlastfoot
L1 & Fine-grained / texture & 0.341 & 0.056 & 0.060 & 0.052 \\
L1 & Environmental shift & 0.377 & 0.092 & 0.067 & 0.043 \\
L1 & Viewpoint / scale & 0.341 & 0.066 & 0.017 & 0.026 \\
L1 & Typographic conflict & 0.360 & 0.059 & 0.043 & 0.020 \\
L2 & Fine-grained / texture & 0.408 & 0.072 & 0.022 & 0.019 \\
L2 & Environmental shift & 0.414 & 0.107 & 0.041 & 0.018 \\
L2 & Viewpoint / scale & 0.351 & 0.066 & 0.021 & -0.009 \\
L2 & Typographic conflict & 0.388 & 0.056 & 0.030 & 0.033 \\
\end{longtable}
\endgroup

\begin{figure}[!htbp]
\centering
\includegraphics[width=0.96\textwidth]{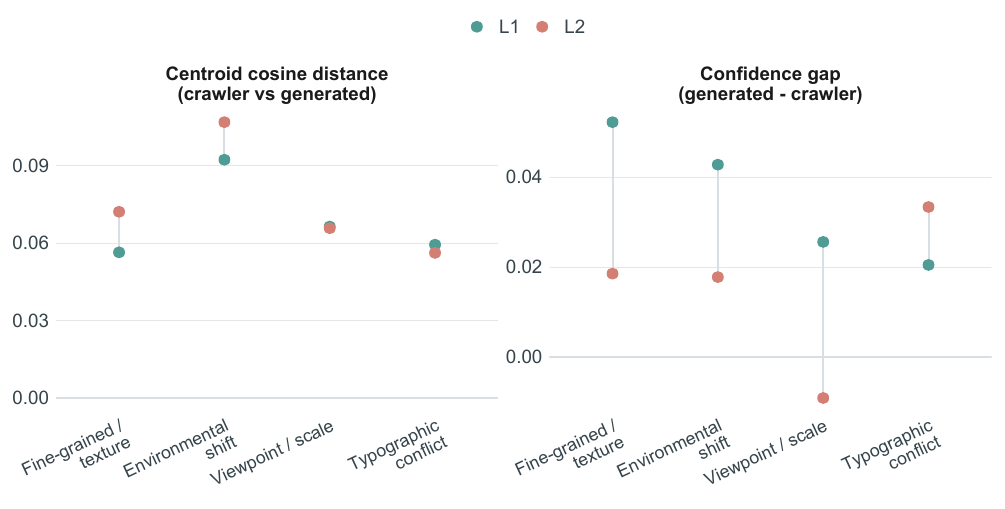}
\caption{Category-level analyses grouped by image source for the four dual-source categories. Signed confidence gaps use generated minus crawler means.}
\label{fig:M2-category-source-diagnostics}
\end{figure}
\FloatBarrier

\subsection{Local Threshold Sensitivity}
\label{app:threshold-sensitivity}
\label{app:gate-threshold-robustness}

For $\delta\in\{-0.05,-0.03,-0.01,+0.01,+0.03,+0.05\}$, a single-threshold setting uses $T_k(\delta)=T_k(1+\delta)$ for one $k\in\{1,2,3\}$ while holding the other thresholds fixed; a joint setting scales all three thresholds. For fixed L2 set $A$ and perturbed set $B$, $J(A,B)=|A\cap B|/|A\cup B|$.

\begin{table}[!htbp]
\centering
\footnotesize
\renewcommand{\arraystretch}{1.06}
\caption{Small local threshold changes and Jaccard summary.}
\label{tab:N1-threshold-jaccard}
\begin{tabularx}{\textwidth}{@{}Xlrrrrr@{}}
\toprule
\AppendixTableHead{Perturbation} & \AppendixTableHead{Range} & \AppendixTableHead{Settings} & \AppendixTableHead{L2 count range} & \AppendixTableHead{Jaccard min} & \AppendixTableHead{Jaccard max} & \AppendixTableHead{Mean} \\
\midrule
Fixed & 0\% & 1 & 14,285--14,285 & 1.000 & 1.000 & 1.000 \\
\shortstack[l]{$T_1$ local\\perturbation} & -5\% to +5\% & 6 & 4,967--16,786 & 0.348 & 0.938 & 0.760 \\
\shortstack[l]{$T_2$ local\\perturbation} & -5\% to +5\% & 6 & 13,214--15,070 & 0.925 & 0.987 & 0.961 \\
\shortstack[l]{$T_3$ local\\perturbation} & -5\% to +5\% & 6 & 10,396--18,560 & 0.728 & 0.944 & 0.844 \\
\shortstack[l]{Joint local\\perturbation} & -5\% to +5\% & 6 & 6,026--14,568 & 0.312 & 0.876 & 0.658 \\
\bottomrule
\end{tabularx}
\end{table}

Single-threshold Jaccard ranges are 0.348--0.938 for $T_1$, 0.925--0.987 for $T_2$, and 0.728--0.944 for $T_3$; joint changes yield 0.312--0.876. This is a local sensitivity analysis and does not establish global threshold optimality.

\begin{figure}[!htbp]
\centering
\includegraphics[width=0.82\textwidth]{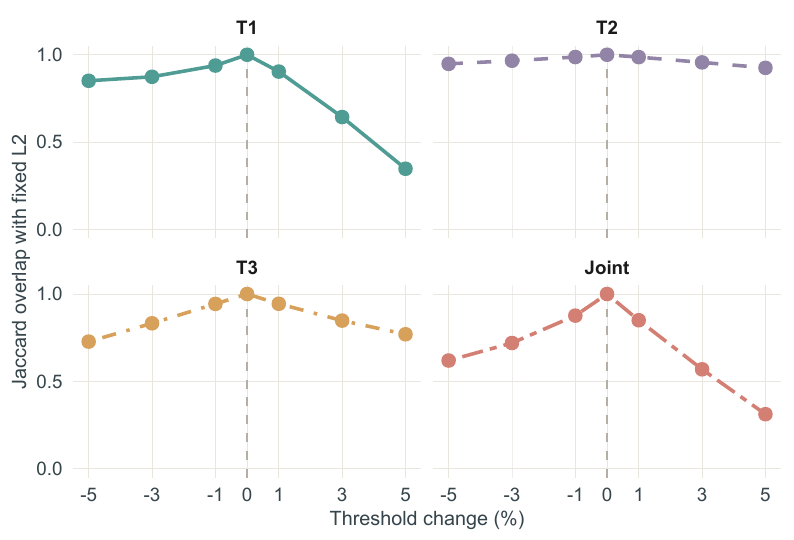}
\caption{L2 overlap under the specified single-threshold and joint local changes. The $0\%$ reference marks fixed L2 membership.}
\label{fig:N1-threshold-jaccard}
\end{figure}
\FloatBarrier

\subsection{Vocabulary Expansion and Fixed-model Diagnostics}
\label{app:fixed-model-diagnostics}
\label{app:wordnet-expansion-check}
\label{app:supplementary-consistency-checks}

The WordNet expansion check after filtering rule 2 measures additional pruning under a broader reference vocabulary. Across categories, additional pruning is 0.39--2.25 percentage points and the overall value is 1.28 percentage points.

\begin{table}[!htbp]
\centering
\footnotesize
\setlength{\tabcolsep}{3.5pt}
\renewcommand{\arraystretch}{1.06}
\caption{WordNet reference-vocabulary expansion check. Additional pruning uses the common number of input images in the calculation.}
\label{tab:O1-wordnet-summary}
\begin{tabular*}{0.96\textwidth}{@{\extracolsep{\fill}}>{\raggedright\arraybackslash}p{3.15cm}rrrrrrr@{}}
\toprule
\AppendixTableHead{Category} & \AppendixTableHead{Input} & \AppendixTableHead{Pre} & \AppendixTableHead{Post} & \AppendixTableHead{\shortstack{Add.\\removed}} & \AppendixTableHead{Pre/input} & \AppendixTableHead{Post/input} & \AppendixTableHead{\shortstack{Add.\\pruning}} \\
\midrule
Time-lagged & 4,299 & 3,747 & 3,728 & 19 & 87.16\% & 86.72\% & 0.44 pp \\
\shortstack[l]{Synthetic /\\fictional} & 3,892 & 3,322 & 3,307 & 15 & 85.35\% & 84.97\% & 0.39 pp \\
\shortstack[l]{Cultural\\blindspots} & 3,550 & 1,919 & 1,839 & 80 & 54.06\% & 51.80\% & 2.25 pp \\
\shortstack[l]{Fine-grained /\\texture} & 3,626 & 1,231 & 1,156 & 75 & 33.95\% & 31.88\% & 2.07 pp \\
\shortstack[l]{Environmental\\shift} & 4,106 & 2,930 & 2,885 & 45 & 71.36\% & 70.26\% & 1.10 pp \\
\shortstack[l]{Viewpoint /\\scale} & 4,386 & 2,447 & 2,410 & 37 & 55.79\% & 54.95\% & 0.84 pp \\
\shortstack[l]{Typographic\\conflict} & 4,651 & 3,563 & 3,469 & 94 & 76.61\% & 74.59\% & 2.02 pp \\
\midrule
Overall & 28,510 & 19,159 & 18,794 & 365 & 67.20\% & 65.92\% & 1.28 pp \\
\bottomrule
\end{tabular*}
\end{table}

For fixed-construction-model logits $\ell_k(x)$ over the 1,000 reference prompts, let
\[
p_k(x)=\frac{\exp(\ell_k(x))}{\sum_{j=1}^{1000}\exp(\ell_j(x))},
\qquad
H(x)=-\sum_{k=1}^{1000}p_k(x)\log\!\left(p_k(x)+10^{-9}\right).
\]
For normalized target-concept encoding $t(a)$, semantic loss is
\[
\mathcal{L}_{\mathrm{sem}}(x,a)=1-z(x)^\top t(a).
\]
The equal-category mean entropy is 3.724 for L1 and 3.757 for L2; mean semantic loss is 0.697 and 0.736, respectively.

\begin{table}[!htbp]
\centering
\footnotesize
\renewcommand{\arraystretch}{1.08}
\caption{Entropy and semantic-loss summary using a mean that gives each of the seven categories equal weight.}
\label{tab:O2-entropy-semantic-loss}
\begin{tabular*}{0.76\textwidth}{@{\extracolsep{\fill}}lrrrr@{}}
\toprule
\AppendixTableHead{Layer} & \AppendixTableHead{Categories} & \AppendixTableHead{Mean entropy} & \AppendixTableHead{Entropy range} & \AppendixTableHead{Mean semantic loss} \\
\midrule
L1 & 7 & 3.724 & 3.494--3.840 & 0.697 \\
L2 & 7 & 3.757 & 3.459--3.975 & 0.736 \\
\bottomrule
\end{tabular*}
\end{table}

\subsection{Semantic Relation to ImageNet-1K}
\label{app:semantic-relation}
\label{app:semantic_relation}

We encode the 560 target concept phrases and 1,000 ImageNet-1K class prompts in the fixed CLIP ViT-B/32 text space and report cosine similarity to the nearest ImageNet prompt. Across all target concept phrases, the mean is 0.706922, median 0.710037, SD 0.086127, Q25 0.652000, Q75 0.765352, minimum 0.418652, and maximum 0.941011.

\begin{table}[!htbp]
\centering
\small
\renewcommand{\arraystretch}{1.08}
\caption{Nearest-ImageNet similarity by fixed concept-activity scope.}
\label{tab:D9-semantic-activity}
\begin{tabular*}{0.62\textwidth}{@{\extracolsep{\fill}}lrrr@{}}
\toprule
\AppendixTableHead{Scope} & \AppendixTableHead{$n$} & \AppendixTableHead{Mean} & \AppendixTableHead{Median} \\
\midrule
All design phrases & 560 & 0.706922 & 0.710037 \\
L1-active phrases & 553 & 0.705152 & 0.708891 \\
L2-active phrases & 519 & 0.706235 & 0.710064 \\
\bottomrule
\end{tabular*}
\end{table}

\begin{figure}[!htbp]
\centering
\includegraphics[width=\textwidth]{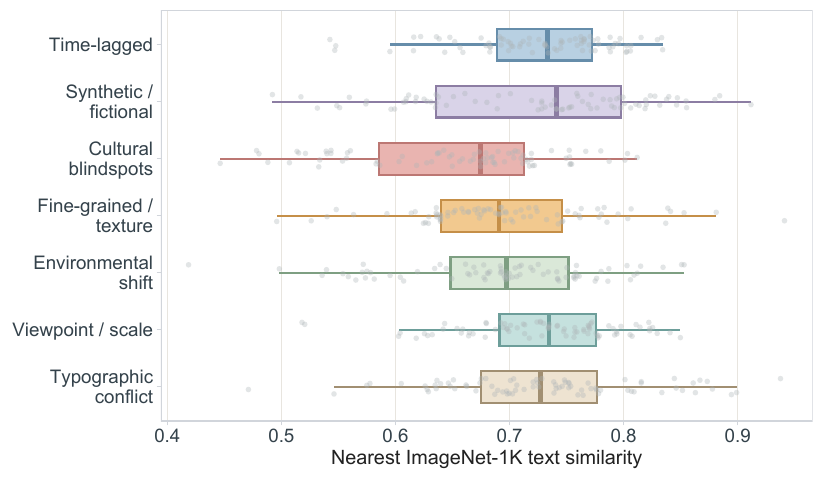}
\caption{Nearest-ImageNet similarity for all 560 target concept phrases in the fixed CLIP ViT-B/32 text space. Boxes summarize each category and light points show individual phrases.}
\label{fig:R1-semantic-reference-relation}
\end{figure}

We reuse the fixed image-level construction scores without new image inference. The lower target-similarity mean in L2 follows a score used by the construction rule and is not an independent validation.

\begin{table}[!htbp]
\centering
\small
\renewcommand{\arraystretch}{1.08}
\caption{Image-level relation between fixed nearest-ID and target-similarity construction scores.}
\label{tab:D10-image-semantic-relation}
\begin{tabular*}{0.82\textwidth}{@{\extracolsep{\fill}}lrrrrr@{}}
\toprule
\AppendixTableHead{Layer} & \AppendixTableHead{$n$} & \AppendixTableHead{Mean nearest-ID} & \AppendixTableHead{Mean target} & \AppendixTableHead{Pearson $r$} & \AppendixTableHead{Spearman $\rho$} \\
\midrule
L1 & 30,752 & 0.254929 & 0.296289 & 0.128880 & 0.115637 \\
L2 & 14,285 & 0.252504 & 0.257976 & 0.067633 & 0.039612 \\
\bottomrule
\end{tabular*}
\end{table}

\begin{figure}[!htbp]
\centering
\includegraphics[width=\textwidth]{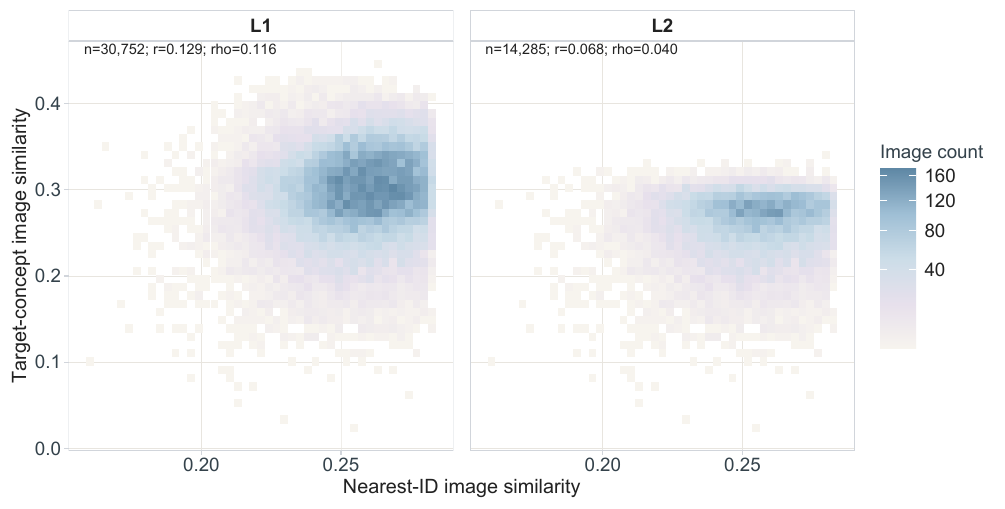}
\caption{Conditioned image-level relation between fixed nearest-ID and target-similarity construction scores. Annotations report full-data descriptive associations.}
\label{fig:R2-semantic-image-relation}
\end{figure}

Prompt wording can affect individual similarity values. ImageNet-1K and one CLIP ViT-B/32 text space are reference-specific rather than universal semantic measures.

\FloatBarrier

\section{Release Boundary and Scope}
\label{app:release-scope}
\label{app:reproducibility-scope}
\setcounter{table}{0}
\setcounter{figure}{0}

\subsection{Intended Use and Release Boundary}
\label{app:release-boundary}
\label{app:reproducibility-release-boundary}

Table~\ref{tab:E1-release-boundary} summarizes the intended use and release boundary of OODConceptBench.

\begin{table}[!htbp]
\centering
\footnotesize
\renewcommand{\arraystretch}{1.10}
\caption{Intended-use and release-boundary summary.}
\label{tab:E1-release-boundary}
\begin{tabularx}{0.92\textwidth}{@{}>{\raggedright\arraybackslash\bfseries}p{0.23\textwidth}>{\raggedright\arraybackslash}X@{}}
\toprule
\AppendixTableHead{Item} & \AppendixTableHead{Specification} \\
\midrule
Intended use & Studying vision--language model recognition boundaries on open-world image--concept pairs. \\
Excluded uses & The benchmark is not a universal safety test or a basis for ranking models across unrelated protocols. \\
Source provenance & Acquisition source is recorded for benchmark images; any public release will retain source information where permitted. \\
Image-level release conditions & Dedicated near-duplicate, sensitive-content, licensing, source, and privacy review, together with generated-content and source-policy documentation. \\
Scope boundary & Evaluation measures target-concept recognition under the stated fixed protocol. \\
\bottomrule
\end{tabularx}
\end{table}

\subsection{Scope and Limitations}
\label{app:limitations}
\label{app:discussion-scope-ethics}
\label{app:discussion-limitations}

The conclusions are limited to the 37 evaluated models. Comparisons across release-year groups, model families, and variants are descriptive because the models differ along multiple dimensions, so observed differences cannot be attributed to a single factor.

The benchmark covers 560 target concept phrases across seven categories rather than the entire open world. Its results characterize recognition on the included image--concept pairs and should not be generalized to every possible concept.

The model-specific $Q_{95,m}$ decision depends on the fixed prompt and reference calibration, while L1/L2 membership depends on the fixed construction model, reference vocabularies, and thresholds. Alternative prompt or calibration choices may change individual measurements; the fixed protocol provides a reproducible basis for comparison.

\FloatBarrier

\end{document}